\documentclass[twoside,11pt]{article}

\usepackage{jmlr2e}

\usepackage{bm} 
\usepackage{xcolor} 
\usepackage{adjustbox} 
\usepackage{tikz} 
\usetikzlibrary{arrows,shapes,positioning,automata}
\usetikzlibrary{calc,decorations.markings}
\usetikzlibrary{decorations.pathreplacing}
\usepackage{amsmath} 
\usepackage{empheq} 
\usepackage[algo2e,ruled,vlined]{algorithm2e} 
\usepackage{booktabs} 
\usepackage{appendix} 

\usepackage{enumitem} 
\usepackage{booktabs}
\usepackage{makecell} 
\usepackage{multirow}
\usepackage{rotating}
\usepackage[most]{tcolorbox} 
\usepackage{float} 
\usepackage{mathtools} 

\usepackage{diagbox} 
\usepackage{wrapfig} 

\jmlrheading{}{}{}{}{}{}{Qi Sun, Hexin Dong, Jiacheng Sun, Zhenguo Li and Bin Dong}

\ShortHeadings{Layer-Parallel Training of Residual Networks}{Sun et al.}
\firstpageno{1}

\begin{document}

\title{Layer-Parallel Training of Residual Networks with Auxiliary-Variable Networks}

\author{\name Qi Sun$^{1,2,}$\thanks{Authors with equal contribution.}$\ ^{,}$\thanks{Corresponding author.} \email qsun$\_$irl@tongji.edu.cn \\
		    \name Hexin Dong$^{3,*}$ \email donghexin@pku.edu.cn \\
		    \name Zewei Chen$^{4}$ \email chen.zewei@huawei.com \\
		    \name Jiacheng Sun$^{4}$ \email sunjiacheng1@huawei.com \\
		    \name Zhenguo Li$^{4}$ \email li.zhenguo@huawei.com \\
		    \name Bin Dong$^{1,3}$ \email dongbin@math.pku.edu.cn \\
       \addr $^{1}$Beijing International Center for Mathematical Research, Peking University, Beijing, China\\
       \addr $^{2}$School of Mathematical Sciences, Tongji University, Shanghai, China \\
       \addr $^{3}$Center for Data Science, Peking University, Beijing, China\\
       \addr $^{4}$Huawei Noah’s Ark Lab}
                     
\editor{ }

\maketitle

\begin{abstract}
Gradient-based methods for the distributed training of residual networks (ResNets) typically require a forward pass of the input data, followed by back-propagating the error gradient to update model parameters, which becomes time-consuming as the network goes deeper. To break the algorithmic locking and exploit synchronous module parallelism in both the forward and backward modes, auxiliary-variable methods have attracted much interest lately but suffer from significant communication overhead and lack of data augmentation. In this work, a novel joint learning framework for training realistic ResNets across multiple compute devices is established by trading off the storage and recomputation of external auxiliary variables. More specifically, the input data of each independent processor is generated from its low-capacity auxiliary network (AuxNet), which permits the use of data augmentation and realizes forward unlocking. The backward passes are then executed in parallel, each with a local loss function that originates from the penalty or augmented Lagrangian (AL) methods. Finally, the proposed AuxNet is employed to reproduce the updated auxiliary variables through an end-to-end training process. We demonstrate the effectiveness of our methods on ResNets and WideResNets across CIFAR-10, CIFAR-100, and ImageNet datasets, achieving speedup over the traditional layer-serial training method while maintaining comparable testing accuracy.
\end{abstract}


\section{Introduction}

With the fast development of computer science and technology, learning with deep neural networks has enjoyed remarkable success across a wide variety of machine learning and scientific computing applications \citep{lecun2015deep,brunton2020machine}, where the backpropagation (BP) algorithm \citep{rumelhart1985learning} is widely adopted to fulfill the learning tasks. However, even with modern Graphical Processing Units (GPUs), the overall training process remains time-consuming, especially when the network is large and distributed across multiple computing devices. 

To improve the training efficiency, various parallelization techniques such as the data-parallelism \citep{iandola2016firecaffe}, model-parallelism \citep{dean2012large}, and a combination of both \citep{paine2013gpu,harlap2018pipedream} have been proposed to reduce the training runtime. Unfortunately, none of these methods could fully overcome the scalability barrier created by the intrinsically serial propagation of data within the network itself \citep{gunther2020layer}, thereby forcing the distributed machines to work synchronously, and hence preventing us from fully leveraging the computing resources. In other words, the forward, backward, and update locking issues \citep{jaderberg2017decoupled} inherited from BP are not addressed by either of the methods above. 

One way of breaking the algorithmic locks and achieving speed-up over the traditional layer-serial training strategies is to apply synthetic gradients to build decoupled neural interfaces \citep{jaderberg2017decoupled}, where the true objective gradients are approximated by auxiliary neural networks so that each layer can be locally updated without performing the full serial BP. However, it fails in training deep convolutional neural networks since the construction of synthetic loss functions has little relation to the target loss function \citep{miyato2017synthetic}. Nevertheless, this pioneering work motivates the continuing research on what later on became known as \textit{local error learning} (see also the early work by \citep{mostafa2018deep} from a biological point of view). Recently, \citep{belilovsky2019greedy,belilovsky2020decoupled,lowe2019putting,pyeon2020sedona,belilovsky2021decoupled} empirically demonstrate that we can apply this kind of method with carefully designed auxiliary neural networks to state-of-the-art networks training on real-world datasets, achieving comparable or even better performance against the benchmark BP approach. However, the loss function of these methods is no longer consistent with the original one, and the local worker can be short-sighted \citep{wang2021revisiting}.

Another kind of method relies on the employment of delayed gradients for updating network parameters and therefore achieves backward unlocking \citep{huo2018decoupled}. However, this design suffers from large memory consumption and weight staleness introduced by the asynchronous backward updates, especially when the network goes deeper \citep{lian2015asynchronous,xu2020acceleration}. To compensate for the gradient delay, feature replay \citep{huo2018training} stores all workers' history inputs and then recomputes the latent feature map before executing the backward operations, which would incur extra computation time hampers the performance for the exposed parallelism. Besides, both methods fail to addresses the forward locking. Hence the acceleration is limited due to Amdahl’s law \citep{gustafson1988reevaluating}. As a potential solution, \citep{xu2020acceleration} suggests overlapping the recomputation with the forward pass to accelerate the training process further, leading to an asynchronous module parallelization. However, the forward dependency of a particular input data still exists.

To exploit synchronous module parallelism in both the forward and backward modes, several algorithms were proposed recently by introducing external auxiliary variables, \textit{e.g.}, the quadratic penalty method \citep{carreira2014distributed,zeng2018global,choromanska2018beyond}, the AL method \citep{taylor2016training,zeng2019convergence,marra2020local} and the proximal method \citep{li2020training} for training fully-connected networks, which can achieve speed-up over the traditional BP algorithm on a single Central Processing Unit (CPU) \citep{li2020training}. However, as shown in \citep{gotmare2018decoupling}, the performances of most of these methods are much worse than that of BP for deep convolutional neural networks. In another recent approach \citep{gunther2020layer,parpas2019predict,kirby2020layer}, based on the similarity of ResNets training to the optimal control of nonlinear systems \citep{weinan2017proposal}, parareal method for solving differential equations is employed to replace the forward and backward passes with iterative multigrid schemes. However, due to its necessity of recording both the state and adjoint variables to perform control updates, the implementation is difficult to integrate with the existing automatic differentiation technologies. Therefore, experiments were conducted on simple ResNets across small datasets rather than state-of-the-art ResNets across modern datasets. So far, to the best of our knowledge, it is still uncertain that whether we can effectively and efficiently apply these layer-parallel training strategies to modern deep ResNets training on real-world datasets.

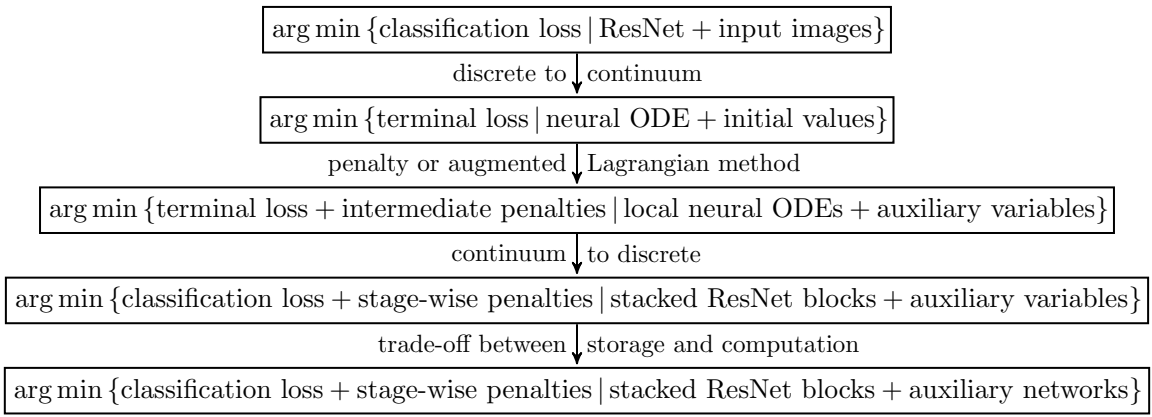
\begin{figure*}[t]
\begin{adjustbox}{max totalsize={\textwidth}{\textheight},center}
\begin{tikzpicture}[shorten >=1pt,auto,node distance=3.5cm,thick,main node/.style={rectangle,draw,font=\normalsize},decoration={brace,mirror,amplitude=7},every text node part/.style={align=center}]
\node[main node] (1) {\parbox{8.57cm}{\centering $\operatorname*{arg\,min} \left\{ \textnormal{classification loss} \, | \, \textnormal{ResNet} + \textnormal{input images} \right\}$}};
\node[main node] (2) [below =0.6cm of 1] {\parbox{8.642cm}{\centering $\operatorname*{arg\,min} \left\{ \textnormal{terminal loss} \, | \, \textnormal{neural ODE} + \textnormal{initial values} \right\}$}};   
\node[main node] (3) [below =0.6cm of 2] {\parbox{14.84cm}{\centering $\operatorname*{arg\,min} \left\{ \textnormal{terminal loss} + \textnormal{intermediate penalties} \, | \, \textnormal{local neural ODEs} + \textnormal{auxiliary variables} \right\}$}};    
\node[main node] (4) [below =0.6cm of 3] {\parbox{15.8cm}{\centering $\operatorname*{arg\,min} \left\{ \textnormal{classification loss} + \textnormal{stage-wise penalties} \, | \, \textnormal{stacked ResNet blocks} + \textnormal{auxiliary variables} \right\}$}};
\node[main node] (5) [below =0.6cm of 4] {\parbox{15.83cm}{\centering $\operatorname*{arg\,min} \left\{ \textnormal{classification loss} + \textnormal{stage-wise penalties} \, | \, \textnormal{stacked ResNet blocks} + \textnormal{auxiliary networks} \right\}$}};

\path[every node/.style={font=\sffamily\small},->,>=stealth']
	(1) edge node[left,midway] {\textcolor{white}{g}\textnormal{discrete to}} node[right,midway] {\textnormal{continuum}\textcolor{white}{g}} (2)
	(2) edge node[left,midway] {\textnormal{penalty or augmented}} node[right,midway] {\textnormal{Lagrangian method}} (3)    
    (3) edge node[left,midway] {\textcolor{white}{g}\textnormal{continuum}} node[right,midway] {\textnormal{to discrete}\textcolor{white}{g}} (4)
    (4) edge node[left,midway] {\textcolor{white}{g}\textnormal{trade-off between}} node[right,midway] {\textnormal{storage and computation}} (5);
\end{tikzpicture}
\end{adjustbox}
\vspace{-0.6cm}
\caption{A diagram describing the motivation of layer-parallel training algorithms from a dynamical systems view.}
\label{fig-big-picture}
\end{figure*}

In this work, a novel layer-parallel training strategy is proposed by using low-capacity auxiliary-variable networks, which can achieve forward, backward, and update unlocking without explicitly storing the external auxiliary variables. As can be seen from \autoref{fig-big-picture}, our method is first motivated by the similarity of training ResNets to the optimal control of neural ODEs, which allows us to understand and break the algorithmic locking in the continuous-time sense. The corresponding discrete dynamical systems then offer us a fully decoupled training scheme, whereas the introduction of auxiliary networks permits the use of data augmentation by trading off storage and recomputation of the external auxiliary variables during training. More specifically, the proposed algorithm is constructed as follows (see also \autoref{fig-big-picture}):
\begin{itemize}
\item[$\bullet$] discrete to continuum: the deep layer limit of ResNet training task coincides with a terminal control problem governed by the so-called neural ordinary differential equations (ODE) \citep{weinan2017proposal,haber2017stable,thorpe2018deep};
\item[$\bullet$] penalty or augmented Lagrangian method: the penalty or AL method \citep{maday2002parareal,nocedal2006numerical} is employed for the approximate solution of the neural ODE-constrained optimal control problem, which breaks the forward, backward, and update locking in the continuous-time sense;
\item[$\bullet$] continuum to discrete: a consistent finite difference scheme \citep{gholami2019anode} is adopted for the discrete-in-time solution of the relaxed problem in the previous step, leading to a layer-parallel training algorithm with external auxiliary variables;
\item[$\bullet$] trade-off between storage and computation: to enable the use of data augmentation for improving model performance, the low-capacity AuxNet is trained to learn the correlation between auxiliary variables by replicating their values at each iteration.
\end{itemize}

The rest of this paper is organized as below. Section 2 is devoted to reviewing the classical BP algorithm and its locking issues, followed by the layer-serial training of deep ResNets from a dynamical systems viewpoint. Then a brief survey of related work on removing these limitations is presented. Next, in section 3, the penalty and AL methods are illustrated to achieve forward, backward, and update unlocking in the continuous-time sense. Its discrete-time counterpart is presented in section 4. Experimental results are reported in section 5 to demonstrate the effectiveness and efficiency of our methods, followed by some concluding remarks and future improvements in section 6.

\section{Background and Related Work} \label{Section-Related-Work}

In this section, we begin by briefly reviewing the most commonly used BP algorithm \citep{hecht1992theory} for training deep ResNets \citep{he2016deep,he2016identity}, followed by a concise description of its locking effects \citep{jaderberg2017decoupled} from the dynamical system view \citep{li2017maximum}. Then a survey of related work on breaking these algorithmic limitations (\textit{i.e.}, the forward, backward, and update lockings) is presented and compared with our current work.

\subsection{Layer-Serial Training From a Dynamical Systems View} \label{section-ResNet}

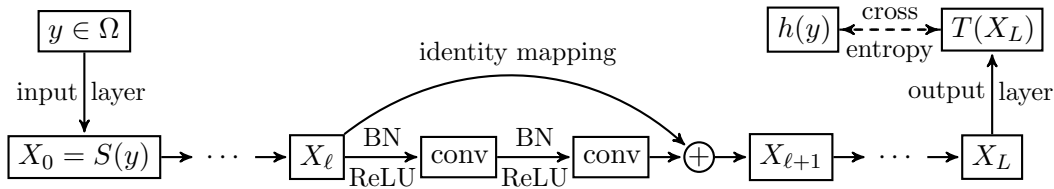
\begin{figure}[t]
\begin{adjustbox}{max totalsize={\textwidth}{\textheight},center}
\begin{tikzpicture}[shorten >=1pt,auto,node distance=2.5cm,thick,main node/.style={rectangle,draw,font=\normalsize},decoration={brace,mirror,amplitude=7}]
\node[main node] (1) {$X_\ell$};
\node[main node] (2) [right = 1cm of 1] {\parbox[c][0.28cm]{0.74cm}{conv} };
\node[main node] (3) [right = 1cm of 2] {\parbox[c][0.28cm]{0.74cm}{conv} };
\node[draw, circle, inner sep=0pt, minimum size=0.21cm, color=black] (4) [right = 0.45cm of 3] {+};
\node[main node] (5) [right = 0.45cm of 4]  {$X_{\ell+1}$};

\node[] (6) [left = 0.45cm of 1]  {$\cdots$};
\node[main node] (7) [left = 0.45cm of 6]  {\parbox[c][0.3cm]{1.76cm}{$X_0=S(y)$} };
\node[] (8) [right = 0.45cm of 5]  {$\cdots$};
\node[main node] (9) [right = 0.45cm of 8]  {$X_{L}$};

\node[main node] (10) [above = 1.1cm of 7] {\parbox[c][0.28cm]{0.96cm}{$y\in \Omega$} };
\node[main node] (11) [above = 1.1cm of 9] {\parbox[c][0.3cm]{1.08cm}{$T(X_L)$} };
\node[main node] (12) [left = 1.35cm of 11] {\parbox[c][0.33cm]{0.68cm}{$h(y)$} };

\path[every node/.style={font=\sffamily\small},->,>=stealth']
    (1) edge node[above,midway] {\textnormal{BN}} node[below,midway] {\textnormal{ReLU}} (2)
    		edge [bend left=40] node[above] {\textnormal{identity mapping}} (4)
    (2) edge node[above,midway] {\textnormal{BN}} node[below,midway] {\textnormal{ReLU}} (3)
    (3) edge node {} (4)
    (4) edge node {} (5)
    (7) edge node {} (6)
    (6) edge node {} (1)
    (5) edge node {} (8)
    (8) edge node {} (9);

\path[every node/.style={font=\sffamily\small},->,>=stealth']
    (10) edge node[left] {\textnormal{input}\!} node[right] {\!\textnormal{layer}} (7)
    (9) edge node[left] {\textnormal{output}\!} node[right] {\!\textnormal{layer}} (11);
\path[every node/.style={font=\sffamily\small},->,>=stealth',dashed]    
    (11) edge node[above,midway] {\textnormal{cross}} node[below,midway] {\textnormal{entropy}} (12)
    (12) edge (11);  
\end{tikzpicture}
\end{adjustbox}
\vspace{-0.5cm}
\caption{A diagram describing the layer-serial training process of a pre-activation ResNet.}
\label{fig-ResNet-architecture}
\end{figure}

We consider the benchmark residual learning framework \citep{he2016deep,he2016identity} that assigns pixels in the raw input image to categories of interest as depicted in \autoref{fig-ResNet-architecture}. Specifically, given a human-labeled database $\{y,h(y)\}_{y\in\Omega}$, the learning task requires solving the optimization problem
\begin{equation}
	\operatorname*{arg\,min}_{\{W_\ell\}_{\ell=0}^{L-1}} \left\{ \varphi(X_L) \, \Big| \, X_0=S(y), \ X_{\ell+1} = X_\ell + F(X_\ell,W_\ell)\ \, \textnormal{for} \ 0\leq \ell\leq L-1 \right\} 
	\label{ResNet-Training-Task}
\end{equation}
where $\displaystyle \varphi(X_L) = \mathbb{E}_{y\in \Omega} \! \Big[ \lVert T(X_L) - h(y) \rVert \Big]$\footnote{Though the population risk is of primary interest, we only have access to the empirical risk in practice. For notational simplicity, we still denote by $\varphi(\cdot)$ the loss function obtained from a particular mini-batch of the entire training dataset throughout this work.} represents the classification loss function, $X_\ell$ the input feature map of the $\ell$-th residual block, $L\in\mathbb{N}_+$ the total number of building blocks, $F$ typically a composition of linear and nonlinear functions as depicted in \autoref{fig-ResNet-architecture}, $W_\ell$ the network parameters to be learned, and $\lVert\cdot\rVert$ a given metric measuring the discrepancy between the model prediction $T(X_L)$ and the ground-truth label $h(y)$ for each input image $y\in\Omega$. The trainable parameters of input and output layers, \textit{i.e.}, $S$ and $T$ in \autoref{fig-ResNet-architecture}, are assumed to be fixed \citep{haber2018learning} for the ease of illustration.

Based on the concept of modified equations \citep{weinan2017proposal} or the variational analysis using $\Gamma$-convergence \citep{thorpe2018deep}, the learning task \eqref{ResNet-Training-Task} can be interpreted as the discretization of a terminal control problem\footnote{For the ease of comparison, we refer, respectively, to problem \eqref{ResNet-Training-Task} and its counterpart \eqref{ODE-Training-Task} as the layer-serial and time-serial training method, and the same is said for their variants in the following sections.} governed by the so-called neural ordinary differential equation \citep{chen2018neural}
\begin{equation}
	\operatorname*{arg\,min}_{\omega_t} \left\{ \varphi(x_1) \, \Big| \, x_0=S(y) , \, dx_t = f(x_t,w_t)dt \ \, \textnormal{for}\ 0<t\leq 1 \right\}.
	\label{ODE-Training-Task}
\end{equation}

Accordingly, the continuous-in-time counterpart of the most commonly used BP algorithm \citep{hecht1992theory} for solving \eqref{ResNet-Training-Task}, namely,
\begin{equation}
	W_\ell \leftarrow W_\ell - \eta \frac{\partial \varphi(X_L)}{\partial X_{\ell+1}}\frac{\partial X_{\ell+1}}{\partial W_\ell}, \qquad 0\leq \ell \leq L-1,
	\label{ResNet-BackPropagation}
\end{equation}
is handled by the adjoint and control equations for finding the extremal of \eqref{ODE-Training-Task} \citep{li2017maximum}, that is, for $0\leq t\leq 1$,
\begin{gather}
dp_t = -p_t \frac{\partial f(x_t,w_t)}{\partial x}dt, \qquad p_1=\frac{\partial \varphi(x_1)}{\partial x},  \label{ODE-Adjoint-Equation}\\
w_t \leftarrow w_t - \eta p_t \frac{\partial f(x_t,w_t)}{\partial w}, \label{ODE-Weights-Update}
\end{gather}
where $\eta>0$ represents the learning rate and $p_t$ the adjoint (or co-state) variable that captures the loss changes with respect to hidden activation layers (see \autoref{appendix-Serial-Train} and \autoref{appendix-Serial-Train-Time} for more details). 

Notably, as the training datasets and modern neural networks are becoming larger and larger, a single machine can no longer meet the memory and computing demands placed on it, leading to a need for distributed training. As an illustrative example, the top figure in \autoref{Table-Related-Works-Fig} shows a partitioning of the deep ResNet \eqref{ResNet-Training-Task} across three machines, where the conventional BP algorithm \eqref{ResNet-BackPropagation} with stochastic gradient descent optimization technique \citep{bottou2010large} is adopted for training. Clearly, such a straightforward implementation would result in several forms of \textit{locking}  \citep{jaderberg2017decoupled}, that is, 
\begin{itemize}
\setlength\itemsep{0.05em}
\item[(i)] forward locking: for each input mini-batch, no worker can process its corresponding incoming data before the previous node in directed forward network have executed;
\item[(ii)] backward locking: no worker can capture the loss changes with respect to its hidden activation layers before the previous node in the backward network have executed;
\item[(iii)] update locking: no worker can update its trainable parameters before all the dependent nodes have executed in both the forward and backward modes;
\end{itemize}
which results in severe under-utilization of computing resources. More specifically, as can be seen from the top figure in \autoref{Table-Related-Works-Fig}, only a single machine is active at any instant of time, leading to a V-shaped cycle with lots of idle machines over the entire training process. 

It is noteworthy that these locking issues for training deep ResNets can be recast as the necessity of first solving the forward-in-time neural ODE in \eqref{ODE-Training-Task} and then the backward-in-time adjoint (or co-state) equation \eqref{ODE-Adjoint-Equation} in order to perform the control updates \eqref{ODE-Weights-Update} \citep{weinan2017proposal,li2017maximum}. This connection not only brings us a dynamical system view of the locking effects but also provides a way to consistently discretize the iterative system (\ref{ODE-Adjoint-Equation}-\ref{ODE-Weights-Update}) for solving the time-serial training problem \eqref{ODE-Training-Task}.

\begin{table}
\setlength{\tabcolsep}{3pt}
\centering
\begin{tabular}{c c c}
\toprule
Method & Training Progress (across 3 workers with 3 mini-batches displayed) \\
\midrule
BP & \vspace*{0.05cm}\raisebox{-0.42\totalheight}{\includegraphics[width=0.75\textwidth]{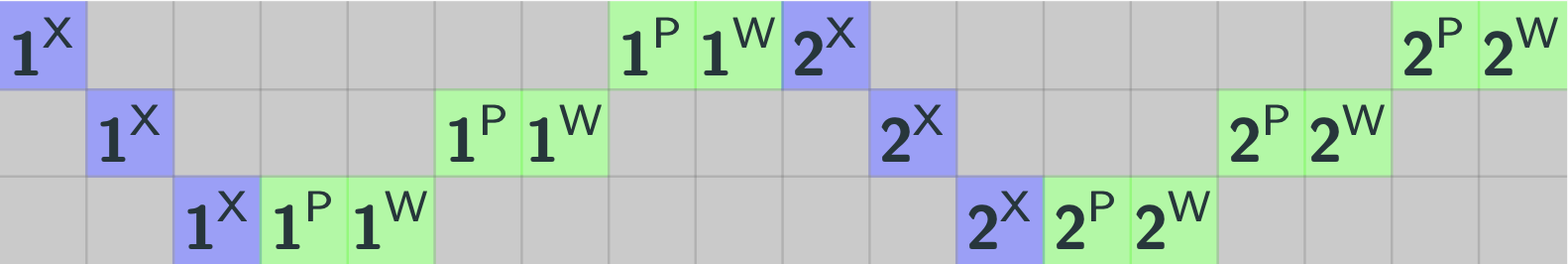}} \\
\midrule
PipeDream & \vspace*{0.05cm}\raisebox{-0.42\totalheight}{\includegraphics[width=0.75\textwidth]{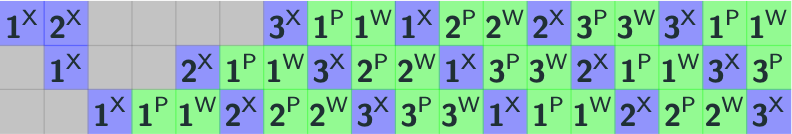}} \\
\midrule
Gpipe & \vspace*{0.05cm}\raisebox{-0.42\totalheight}{\includegraphics[width=0.75\textwidth]{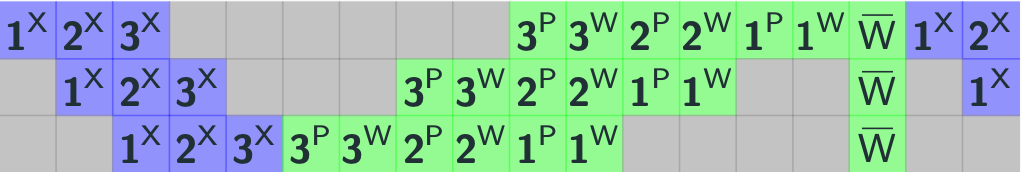}} \\
\midrule
DDG & \vspace*{0.05cm}\raisebox{-0.42\totalheight}{\includegraphics[width=0.75\textwidth]{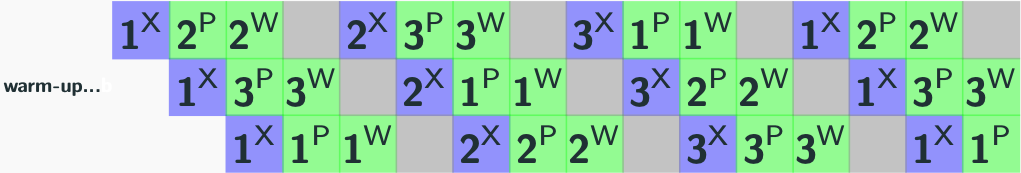}} \\
\midrule
FR & \vspace*{0.05cm}\raisebox{-0.42\totalheight}{\includegraphics[width=0.75\textwidth]{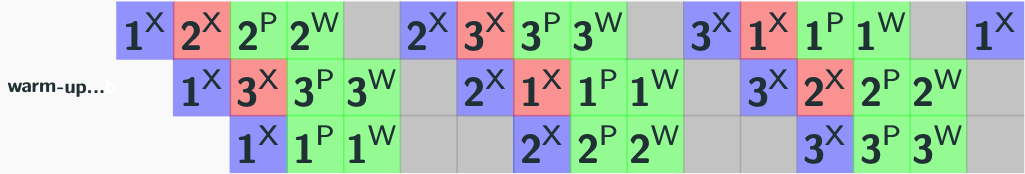}} \\
\midrule
Sync-DGL & \vspace*{0.05cm}\raisebox{-0.42\totalheight}{\includegraphics[width=0.75\textwidth]{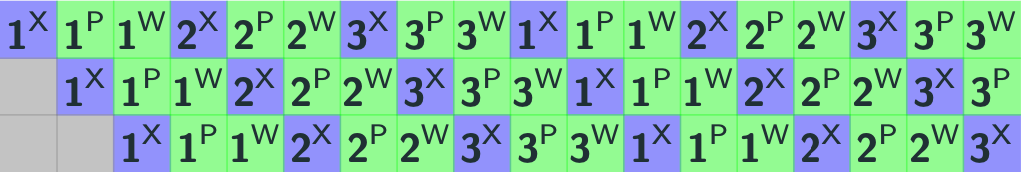}} \\
\midrule
Async-DGL & \vspace*{0.05cm}\raisebox{-0.42\totalheight}{\includegraphics[width=0.75\textwidth]{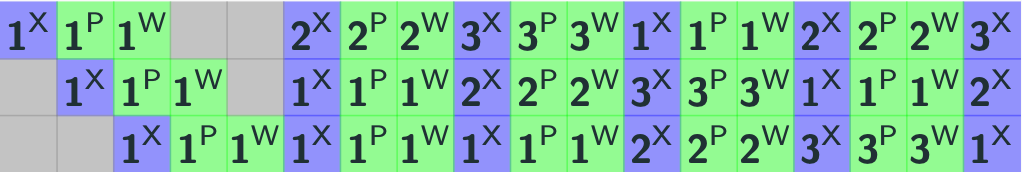}} \\
\midrule
Our Method & \vspace*{0.05cm}\raisebox{-0.42\totalheight}{\includegraphics[width=0.75\textwidth]{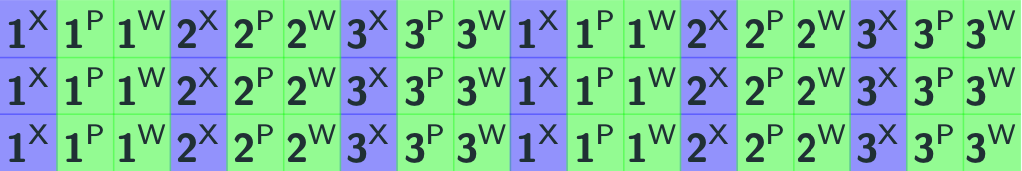}} \\
\bottomrule
\bottomrule
\vspace*{-0.37cm}\\
\makecell{Notation \\ Description} & \vspace*{0.05cm}\raisebox{-0.42\totalheight}{\includegraphics[width=0.75\textwidth]{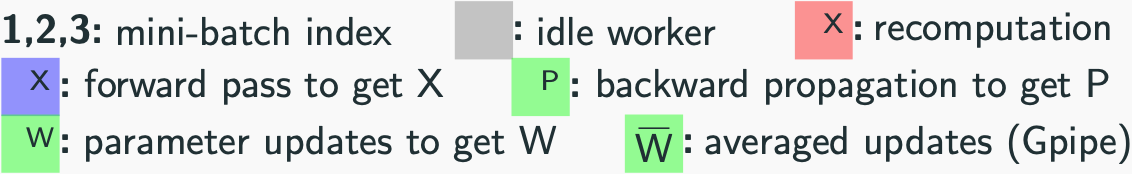}} \\
\bottomrule
\end{tabular}
\caption{Timelines of different strategies to accelerate the neural network training process (runtime and per-worker utilization are shown along the $x$- and $y$-axes). For each worker, the backward pass is assumed to take twice as long as the forward one.}
\label{Table-Related-Works-Fig}
\end{table}

\subsection{Related Works}

For models that are too large to store in a single worker's memory or cache, some representative works on breaking the forward, backward, and update locking issues are briefly reviewed and summarized in this section, as illustrated in \autoref{Table-Related-Works-Fig} and \autoref{Table-Related-Works}. 

\textit{Backpropagation:} Due to the necessity of first executing the forward pass of \eqref{ResNet-Training-Task} and then the backward gradient propagation \eqref{ResNet-Gradient-Propagation} in order to perform parameter updates \eqref{ResNet-Parameter-Updates} (also known as the forward, backward, and update locking, respectively) at each iteration step, the conventional BP algorithm \citep{hecht1992theory} for training deep neural networks forces the distributed machines to work in a synchronous fashion (see the top figure in \autoref{Table-Related-Works-Fig}), thereby compromising the efficiency of computing resources.

\textit{Pipeline Parallelism:} To accelerate the distributed training process, PipeDream \citep{harlap2018pipedream,narayanan2019pipedream,narayanan2021memory} and GPipe \citep{huang2019gpipe} propose the \textit{pipelined} model parallelism so that multiple input data can be pushed through all the available workers in a sequential order. To be specific, PipeDream pipelines the execution of forward passes and intersperses them with BPs in an attempt to minimize the processor idle time. However, such an asynchronous approach introduces inconsistent parameter updates between the forward and backward passes of a particular input (see \autoref{Table-Related-Works-Fig} and \eqref{ResNet-Parameter-Updates}), requiring more effort to compute the gradient accurately \citep{narayanan2021memory}. On the other hand, GPipe divides the input mini-batch into smaller micro-batches (not relabeled in \autoref{Table-Related-Works-Fig}) and updates parameter after accumulating all the micro-batch gradients, which can be regarded as a combination of model parallelism and data parallelism.

Unfortunately, the locking issues inherited from the forward-backward propagation of data across all the network layers are not addressed by the pipeline parallelism. Therefore, exploring dimensions beyond the data and model parallelism may potentially further accelerate the network training.

\textit{Delayed Gradients:} Note that the backward pass takes roughly twice as long as the forward one, which motivates the pursuit of removing the backward locking. \citep{huo2018decoupled} proposes decoupled parallel backpropagation (DDG), achieving backward unlocking by applying delayed gradients for parameter updates. However, this design is known to suffer from large memory consumption and weight staleness (see \autoref{Table-Related-Works-Fig}) introduced by asynchronous backward updates, especially when the network goes further deeper \citep{lian2015asynchronous,xu2020acceleration}. To compensate the gradient delay (see \autoref{Table-Related-Works-Fig}), feature replay (FR) \citep{huo2018training} stores the history inputs of all workers and recompute the hidden activations before executing the backward operations, which would incur additional computation time. Besides, neither DDG nor FR addresses the issue of forward locking, hence the acceleration is limited due to the Amdahl's law \citep{gustafson1988reevaluating}. \citep{xu2020acceleration} suggests to overlap the recomputation with the forward pass to further reduce the runtime, however, the forward dependency of a particular input data still exists.

\textit{Local Error Learning:} Another effective way of removing the backward locking is to artificially build local loss functions so that the trainable parameters of each individual worker can be locally updated without performing the full layer-serial BP. One of the pioneer works is the decoupled neural interface \citep{jaderberg2017decoupled}, which employs auxiliary neural networks to generate synthetic gradients for the decoupled backward pass.  However, it fails in training deep convolutional neural networks since the auxiliary local loss function has little relation to the target objective function \citep{miyato2017synthetic}. To further reduce the discrepancy between the synthetic and exact gradients, \citep{mostafa2018deep} suggests to use local classifiers with cross-entropy loss for the backward propagation, while a similarity measure combined with the local classifier is introduced in \citep{nokland2019training}. Recently, \citep{belilovsky2019greedy,belilovsky2020decoupled,lowe2019putting,pyeon2020sedona} empirically demonstrate that this kind of method with carefully designed auxiliary networks can be applied to modern deep networks across real-world datasets, achieving comparable or even better performance against the standard BP. More specifically, as can be seen from \autoref{Table-Related-Works-Fig}, the synchronous decoupled greedy learning (Sync-DGL) \citep{belilovsky2020decoupled} attaches each worker with a small auxiliary network that generates its own gradients for BP, which allows achieving backward unlocking. Moreover, by making use of a replay buffer for each worker, forward unlocking can be achieved but in an asynchronous fashion (see the Async-DGL in \autoref{Table-Related-Works-Fig}).

However, the objective function of local error learning methods is no longer consistent with the original one \eqref{ResNet-Training-Task}, or in other words, the actual problem has been changed to another one by these methods. In this way, perhaps it is more appropriate to say that these methods themselves do not suffer from the forward, backward, and update locking, rather than breaking the locking issues of the baseline model \eqref{ResNet-Training-Task}. Another sacrifice that local error learning has to make is the introduction of auxiliary networks with extra trainable parameters, forcing the workers to be short-sighted and learn features that only benefit local layers \citep{wang2021revisiting}.

\textit{Auxiliary Variable Methods:} To exploit model parallelism in both the forward and backward modes, one can also consider the use of external auxiliary variables for distributed training, \textit{e.g.}, the quadratic penalty method \citep{choromanska2018beyond,zeng2018global,carreira2014distributed}, the augmented Lagrangian method \citep{taylor2016training,zeng2019convergence,marra2020local}, and the proximal method \citep{li2020training} for training fully-connected networks, which can achieve speed-up over the traditional layer-serial training strategy on a single Central Processing Unit (CPU) \citep{li2020training}. However, as shown in \citep{gotmare2018decoupling}, the performances of most of these methods are much worse than that of BP for deep convolutional neural networks. On the other hand, based on the similarity of ResNets training to the optimal control of nonlinear systems \citep{weinan2017proposal}, the parareal method for solving differential equations is employed to replace the conventional forward-backward propagation with iterative multigrid schemes \citep{gunther2020layer,parpas2019predict,kirby2020layer}. Although the locking issues can be resolved, the implementation is complicated and difficult to integrate with the existing library technologies such as BP and automatic differentiation. Therefore, experiments were conducted on the simple ResNets across small datasets \citep{kirby2020layer}, rather than the state-of-the-art ResNets across larger datasets. 

So far, to the best of our knowledge, it is still uncertain that whether these auxiliary variable methods can be effectively and efficiently applied to modern deep networks across real-world datasets, which motivates us to combine the benefits of all these technologies and develop a novel layer-parallel training method using auxiliary variable networks\footnote{To clarify the differences between model-parallel training of fully-connected networks \citep{zeng2019convergence} and ResNets, we refer the readers to \autoref{fig-big-picture} for technical details.}. To be specific, as can be seen from \autoref{Table-Related-Works-Fig}, the input data of each worker is induced from its corresponding auxiliary variable networks, which permits the use of data augmentation at a low cost and achieves forward unlocking. Then the backward passes are executed synchronously in parallel, each with a local loss function that originates from the penalty or AL methods (see \autoref{fig-big-picture}).

\begin{table}[h]
\setlength{\tabcolsep}{4.pt}
\centering
\begin{tabular}{c c c c c c c c}
\toprule
\multirow{2}{*}[-0.27em]{Method} & \multirow{2}{*}[-0.27em]{\!\!\makecell{Model \\ Solution}} & \multicolumn{3}{c}{Module Unlocking} & \multirow{2}{*}[-0.27em]{\makecell{Forward \\ Independency}} & \multicolumn{2}{c}{Auxiliary Request} \\
\cmidrule(lr){3-5}
\cmidrule(lr){7-8}
&   &  Forward  &  Backward  &  Update  & &  Memory  &  Network  \\
\midrule
BP & $\surd$ & $\times$ & $\times$ & $\times$ & $\times$ & $\times$ & $\times$ \\
\midrule
PipeDream & $\surd$ & $\times$ & $\times$ & $\times$ & $\times$ & $\times$ & $\times$ \\
\midrule
Gpipe & $\surd$ & $\times$ & $\times$ & $\times$  & $\times$ & $\surd$ & $\times$ \\
\midrule
DDG & $\surd$ & $\times$ & $\surd$ & $\surd$ & $\times$ & $\surd$ & $\times$ \\
\midrule
FR & $\surd$ & $\times$ & $\surd$ & $\surd$ & $\times$ & $\surd$ & $\times$ \\
\midrule
\parbox{1.9cm}{\! Sync-DGL \!} & $\times$ & $\times$ & $\surd$ & $\surd$ & $\times$ & $\times$ & $\surd$ \\
\midrule
\parbox{2.1cm}{\! Async-DGL \!} & $\times$ & $\surd$ & $\surd$ & $\surd$ & $\times$ & $\surd$ & $\surd$ \\
\midrule
\parbox{2.1cm}{\! Our Method \!} & $\surd$ & $\surd$ & $\surd$ & $\surd$ & $\surd$ & $\times$ & $\surd$ \\
\bottomrule
\end{tabular}
\caption{Comparison of different model-parallel training methods in terms of model solution, module unlocking, forward independency and auxiliary request.}
\label{Table-Related-Works}
\end{table}

To sum up, \autoref{Table-Related-Works} compares our method with other representative works in terms of several criteria, all of which can not only decouple the traditional layer-serial training process but also show potential to match the performance of a standard BP in real-world applications. In either case, the training strategy should be consistent with the default baseline model \eqref{ResNet-Training-Task} for fair comparison, which we refer to as \textit{model solution} in \autoref{Table-Related-Works}. It is noteworthy that although the forward unlocking can be achieved by allowing the workers to operate asynchronously on different mini-batches (\textit{e.g.}, the Async-DGL in \autoref{Table-Related-Works-Fig}), the forward pass of a particular mini-batch still remains sequential across multiple compute devices. For the ease of illustration, removing such a specific forward locking is called \textit{forward independency} in \autoref{Table-Related-Works}. On the other hand, backward unlocking would permit the decoupled BP running in parallel once the local forward pass is completed, while update unlocking would allow updates of model parameters before all subsequent modules complete executing the forward propagation. Besides, the common sacrifice for achieving module-parallelism is the introduction of auxiliary memory footprints or neural networks, which is summarized in the last two columns of \autoref{Table-Related-Works}. Clearly, our method surpasses the other algorithms in terms of the first five criteria, but comes at the cost of incorporating auxiliary networks. Fortunately, the extra trainable parameters imposed by these auxiliary networks are of small size, which is empirically demonstrated in \autoref{sectioin-experiment}.

\section{Time-Parallel Training} \label{section-parareal-terminal-control}

Based on the analysis of forward, backward, and update locking effects in \autoref{Section-Related-Work}, we are now ready to break these algorithmic limitations in the continuous-time sense, \textit{i.e.}, the second path in \autoref{fig-big-picture}. More specifically, to avoid the necessity of solving forward-backward differential equations for performing control updates at each iteration step (see \eqref{ResNet-ODE-KKT-System} in \autoref{appendix-Serial-Train-Time}), constraint violations through the use of penalty and AL methods \citep{nocedal2006numerical,maday2002parareal,carraro2015multiple} are introduced in this section, which also paves the way for increasing the concurrency of the network training process.

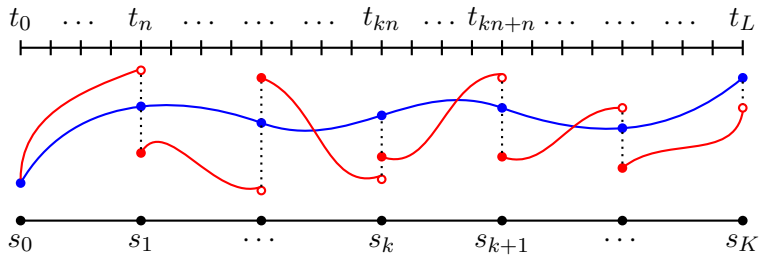
\begin{figure*}[t]
\begin{adjustbox}{max totalsize={\textwidth}{\textheight},center}
\begin{tikzpicture}[xscale=1]
\draw [black] [thick] (0.8,-0.2) -- (10.4,-0.2);
\foreach \x in {0,...,24}
      \draw [black] [thick] (0.8+0.4*\x,-.3) -- (0.8+0.4*\x,-.1);
\node[above] at (0.8,-.1) {$t_0$};
\node[above] at (1.6,-.1) {$\cdots$};
\node[above] at (2.4,-.1) {$t_n$};
\node[above] at (3.2,-.1) {$\cdots$};
\node[above] at (4,-.1) {$\cdots$};
\node[above] at (4.8,-.1) {$\cdots$};
\node[above] at (5.6,-.1) {$t_{kn}$};
\node[above] at (6.4,-.1) {$\cdots$};
\node[above] at (7.2,-.13) {$t_{kn+n}$};
\node[above] at (8,-.1) {$\cdots$};
\node[above] at (8.8,-.1) {$\cdots$};
\node[above] at (9.6,-.1) {$\cdots$};
\node[above] at (10.4,-.1) {$t_L$};

\draw [-,thick, blue] (0.8,-2) to [out=60,in=160] (4,-1.2)
to [out=340,in=160] (7.2,-1) to [out=340,in=220] (10.4,-0.6);

\draw [black] [thick] (0.8,-2.5) -- (10.4,-2.5);
\foreach \x in {0,...,6}
      \draw [fill=black] [black] [thick] (0.8+1.6*\x,-2.5) circle [radius=.05];
\node[below] at (0.8,.-2.55) {$s_0$};
\node[below] at (2.4,.-2.55) {$s_1$};
\node[below] at (4,.-2.55) {$\cdots$};
\node[below] at (5.6,.-2.55) {$s_k$};
\node[below] at (7.2,.-2.55) {$s_{k+1}$};
\node[below] at (8.8,-2.55) {$\cdots$};
\node[below] at (10.4,.-2.55) {$s_K$};

\draw [fill=blue] [blue] [thick] (0.8,-2) circle [radius=.05];
\draw [fill=red] [blue] [thick] (2.4,-0.98) circle [radius=.05];
\draw [red] [thick] (2.4,.-0.5) circle [radius=.05];
\draw [-,thick,red] (0.8,-1.95) to [out=90,in=200] (2.35,.-0.5);

\draw [fill=blue] [blue] [thick] (4,.-1.2) circle [radius=.05];
\draw [fill=red] [red] [thick] (2.4,-1.6) circle [radius=.05];
\draw [red] [thick] (4,.-2.1) circle [radius=.05];
\draw [-,thick,red] (2.4,-1.6) to [out=60,in=200] (4,.-2.05);
\draw [black] [dotted,thick] (2.4,-0.56) -- (2.4,-0.92) ;
\draw [black] [dotted,thick] (2.4,-1.03) -- (2.4,-1.55) ;
\draw [black] [dotted,thick] (4,-0.55) -- (4,-1.15) ;
\draw [black] [dotted,thick] (4,-2.05) -- (4,-1.23) ;

\draw [fill=red] [red] [thick] (4,-0.6) circle [radius=.05];
\draw [red] [thick] (5.6,.-1.95) circle [radius=.05];
\draw [-,thick,red] (4,-0.6) to [out=340,in=200] (5.6,.-1.9);

\draw [fill=blue] [blue] [thick] (5.6,.-1.1) circle [radius=.05];
\draw [fill=blue] [blue] [thick] (7.2,.-1) circle [radius=.05];
\draw [fill=red] [red] [thick] (5.6,.-1.65) circle [radius=.05];
\draw [-,thick,red] (5.6,.-1.65) to [out=340,in=180] (7.2,.-0.55);
\draw [red] [thick] (7.2,.-0.6) circle [radius=.05];
\draw [black] [dotted,thick] (5.6,.-1.6) -- (5.6,.-1.15) ;
\draw [black] [dotted,thick] (5.6,.-1.9) -- (5.6,.-1.7) ;

\draw [fill=blue] [blue] [thick] (8.8,.-1.27) circle [radius=.05];
\draw [fill=red] [red] [thick] (7.2,.-1.65) circle [radius=.05];
\draw [red] [thick] (8.8,.-1) circle [radius=.05];
\draw [-,thick,red] (7.2,.-1.65) to [out=340,in=180] (8.75,.-1);
\draw [black] [dotted,thick] (7.2,.-1.6) -- (7.2,.-1.05);
\draw [black] [dotted,thick] (7.2,.-0.65) -- (7.2,.-0.95) ;

\draw [fill=blue] [blue] [thick] (10.4,.-0.6) circle [radius=.05];
\draw [fill=red] [red] [thick] (8.8,.-1.8) circle [radius=.05];
\draw [red] [thick] (10.4,.-1) circle [radius=.05];
\draw [-,thick,red] (8.8,.-1.8) to [out=40,in=260] (10.4,.-1.05);
\draw [black] [dotted,thick] (8.8,.-1.32) -- (8.8,.-1.8);
\draw [black] [dotted,thick] (8.8,.-1.06) -- (8.8,.-1.22);
\draw [black] [dotted,thick] (10.4,.-0.65) -- (10.4,.-0.95);
\end{tikzpicture}
\end{adjustbox}
\caption{Contrary to the trajectory of neural ODE (blue line), introducing external auxiliary variables at certain moments (solid red dots) enables a time-parallel computation of the state (red lines), adjoint, and control variables. However, to approximately recover the original dynamic, violation of equality constraints (mismatch between the solid and hollow red dots) should be penalized in the loss function.}
\label{fig-multifidelity-forward-propagation}
\end{figure*}

\subsection{Forward Pass with Auxiliary Variables} \label{section-parareal-terminal-control-FP}

To employ $K\in\mathbb{N}_+$ independent processors for the forward-in-time evolution of the neural ODE in \eqref{ODE-Training-Task}, we introduce a coarse partition of $[0,1]$ into several disjoint intervals as shown in \autoref{fig-multifidelity-forward-propagation}, that is,
\begin{equation*}
    0 = s_0  < \ldots < s_k < s_{k+1} < \ldots < s_{K}=1.
\end{equation*}

Now we are ready to define the piecewise state variables $\{x_t^k\}_{k=0}^{K-1}$ such that the underlying dynamic\footnote{We use the superscript $k$ in $x_t^k$, $p_t^k$, and $w_t^k$ to distinguish the disjoint interval $[s_k,s_{k+1}]$ that the state, adjoint, and control variables belonging to.} evolves according to 
\begin{equation}
	x^k_{s_k^+}=\lambda_k, \qquad  d x^k_t = f(x^k_t,w^k_t)dt\ \ \ \textnormal{on}\ (s_k,s_{k+1}],
	\label{Parallel-ODE-State-Equation}
\end{equation}
\textit{i.e.}, the continuous-time forward pass  that originates from external auxiliary variable $\lambda_k$ and with control variable $w_t^k$. Here, and in what follows, $x^k_{s_k^+}$ and $x^k_{s_k^-}$ refer to the right and left limits of the possibly discontinuous function $x^k_t$ at the time interface $t=s_k$. 

Clearly, for any $t\in[s_k,s_{k+1}]$ where $0\leq k\leq K-1$, the state solution of time-serial training method \eqref{ODE-Training-Task} satisfies $x_t = x^k_t$ if and only if
\begin{equation*}
	w_t^k = w_t |_{(s_k,s_{k+1})} \qquad \textnormal{and} \qquad \lambda_k = x_{s_k^+},
\end{equation*}
or by defining $x^{-1}_{s_{0}^-}=x_0$, we arrive at an equivalent form of the above conditions
\begin{equation*}
	w_t^k = w_t |_{(s_k,s_{k+1})} \qquad \textnormal{and} \qquad \lambda_k = x^{k-1}_{s_{k}^-}.
\end{equation*}
As an immediate result, the optimization problem \eqref{ODE-Training-Task} can be reformulated as
\begin{equation}
\begin{gathered}
	\operatorname*{arg\,min}_{\{w^k_t\}_{k=0}^{K-1}} \left\{ \varphi(x^{K-1}_{s_K^-})  \, \Big| \, x^{k-1}_{s_{k}^-}=\lambda_k \ \ \text{and}\ \ x^k_{s_k^+}=\lambda_k, \ \, d x^k_t = f(x^k_t,w^k_t)dt\ \ \textnormal{on}\ (s_k,s_{k+1}] \right\}
\end{gathered}
	\label{ODE-Training-Task-ReWritten}
\end{equation}
which offers the possibility of parallelizing the piecewise neural ODEs \eqref{Parallel-ODE-State-Equation} by removing the other constraint. To be specific, the exact connection between adjacent sub-intervals can be loosened by incorporating external auxiliary variables (see \autoref{fig-multifidelity-forward-propagation}), which inspires us to relax the first constraint in \eqref{ODE-Training-Task-ReWritten} and add penalties to the objective loss function, \textit{i.e.},
\begin{equation}
	\operatorname*{arg\,min}_{\{w_t^k,\lambda_k\}_{k=0}^{K-1}} \bigg\{ \varphi(x^{K-1}_{s_K^-}) + \beta \sum_{k=0}^{K-1} \psi(\lambda_k,x^{k-1}_{s_{k}^-}) \, \Big| \, x^k_{s_k^+}=\lambda_k, \ \, d x^k_t = f(x^k_t,w^k_t)dt\ \ \textnormal{on}\ (s_k,s_{k+1}] \bigg\}
	\label{Parallel-ODE-Training-Task}
\end{equation}
where $\beta>0$ is a scalar constant and $\psi(\lambda,x)= \| \lambda - x\|_{\ell_2}^2$ the penalty function. 

\subsection{Penalty Method} \label{section-penalty-method}

Note that, by definition, $\psi(\lambda_0,x^{-1}_{s_0^-}) = \psi(x_0,x_0) = 0$. The Lagrange functional\footnote{The derivation follows directly from the AL method discussed in the next section and hence is omitted.} \citep{liberzon2011calculus} associated with the optimization problem \eqref{Parallel-ODE-Training-Task}, \textit{i.e.},
\begin{equation*}
	\mathcal{L}_\textnormal{P}( x_t^k, p_t^k, w^k_t, \lambda_k ) = \varphi(x^{K-1}_{s_K^-})  + \sum_{k=0}^{K-1} \left( \beta \psi(\lambda_k,x^{k-1}_{s_{k}^-}) + \int_{s_k}^{s_{k+1}} p^k_t \big( f(x_t^k,w^k_t) - \dot{x}_t^k \big)dt \right)
\end{equation*}
implies that the adjoint variable $p_t^k$ satisfies a backward-in-time differential equation
\begingroup
\renewcommand*{\arraystretch}{2.5}
\vspace{-0.22cm}
\begin{equation}
\begin{array}{l}
	\displaystyle dp_t^{k} = - p_t^{k}\frac{\partial f(x_t^{k},w^k_t)}{\partial x} dt \ \ \ \textnormal{on}\ [s_k,s_{k+1}),\\
	\displaystyle p^k_{s_{k+1}^-} = \left(1-\delta_{k,K-1}\right) \beta \frac{\partial \psi(\lambda_{k+1},x^k_{s_{k+1}^-})}{\partial x} +\delta_{k,K-1} \, \frac{\partial \varphi(x^{k}_{s_{k+1}^-})}{\partial x},
\end{array}
	\label{Parallel-ODE-Adjoint-Equation}
\end{equation}
\endgroup
for any $0\leq k\leq K-1$. Here, the notation $\delta_{k,K-1}$ (or $\delta$ for short) indicates the Kronecker Delta function throughout this work.

Similar to \eqref{ODE-Weights-Update}, the update rule for the piecewise control variables of \eqref{Parallel-ODE-Training-Task} follows from
\begin{equation}
	w_t^k \leftarrow w_t^k - \eta p_t^{k} \frac{\partial f(x_t^k,w^k_t)}{\partial w} \ \ \ \textnormal{on} \ [s_k,s_{k+1}],
	\label{Parallel-ODE-Weights-Update}
\end{equation}
where $0\leq k\leq K-1$, and various gradient-based approaches \citep{gotschel2019efficient,nocedal2006numerical} can be applied for the correction of auxiliary variables, \textit{e.g.},
\begin{equation}
	\lambda_0 \equiv x_0\ \ \ \textnormal{and}\ \ \ \lambda_k \leftarrow \lambda_k - \eta \left( \beta \frac{\partial \psi(\lambda_k,x^{k-1}_{s_{k}^-})}{\partial \lambda} + p^k_{s_k^+} \right)\ \ \textnormal{for}\ 1\leq k\leq K-1.
	\label{Parallel-ODE-Slack-Variables-Update}
\end{equation}

To sum up, the penalty method for approximating the time-serial training procedure \eqref{ODE-Training-Task} at each iteration consists of 
\begin{tcolorbox}[standard jigsaw,opacityback=0,]
\vspace{-0.52cm}					
\begin{empheq}{align*}
	\bullet \ \textnormal{decoupled local operations}\ \eqref{Parallel-ODE-State-Equation}, \eqref{Parallel-ODE-Adjoint-Equation}, \eqref{Parallel-ODE-Weights-Update} \ \qquad \ \bullet \ \textnormal{global communication}\ \eqref{Parallel-ODE-Slack-Variables-Update}
\end{empheq}
\end{tcolorbox}
\noindent which breaks the forward, backward, and update locking in the continuous-time sense.

In particular, by choosing the quadratic penalty function $\psi(\lambda,x) = \lVert \lambda-x \rVert_{\ell_2}^2$, it can be deduced from \eqref{Parallel-ODE-Slack-Variables-Update} that the constraint violations of the approximate minimizer to \eqref{Parallel-ODE-Training-Task} satisfy
\begin{equation}
	  \lambda_0 - x^{-1}_{s_{0}^-} = 0\ \ \ \textnormal{and}\ \ \ \lambda_k - x^{k-1}_{s_{k}^-}  \approx - \frac{1}{2\beta}p^k_{s_k^+}\ \ \textnormal{for}\ 1\leq k\leq K-1,
	  \label{Constraint-Violation-Penalty}
\end{equation}
which implies that a relatively large penalty coefficient $\beta$ is needed in order to force the minimizer of \eqref{Parallel-ODE-Training-Task} close to the feasible region of our original problem \eqref{ODE-Training-Task}. Such a quadratic penalty method has been extensively employed due to its simplicity and intuitive appeal \citep{gotmare2018decoupling,choromanska2018beyond}, however, it suffers from well-known limitations such as numerical ill-conditioning due to the large penalty coefficients \citep{nocedal2006numerical}, memory issue of storing all the external auxiliary variables, and poor quality for feature extraction \citep{gotmare2018decoupling}. Alternatively, the nonsmooth exact penalty method, \textit{e.g.}, $\psi(\lambda,x) = \lVert \lambda-x \rVert_{\ell_1}$ or $\lVert \lambda-x \rVert_{\ell_\infty}$, can often find a solution by performing a single unconstrained minimization, but the non-smoothness may create complications \citep{nocedal2006numerical}. In addition, the schedule of penalty coefficient is often different for different datasets, hence a wise strategy for choosing penalty function and coefficient is of crucial importance to the practical performance.

\subsection{Augmented Lagrangian Method}

To make the approximate solution of \eqref{Parallel-ODE-Training-Task} nearly satisfy the dynamic \eqref{ODE-Training-Task-ReWritten} even for moderate values of penalty coefficient $\beta$, we consider the AL functional
\begin{equation}
	\mathcal{L}_{\textnormal{AL}}( x_t^k, p_t^k, w^k_t, \lambda_k, \kappa_k ) = \mathcal{L}_{\textnormal{P}}( x_t^k, p_t^k, w^k_t, \lambda_k ) - \sum_{k=0}^{K-1} \kappa_k( \lambda_k - x^{k-1}_{s_{k}^-} )
	\label{AL-functional}
\end{equation}
where $\kappa_k$ denotes an explicit Lagrange multiplier associated with the $k$-th continuity-constraint of \eqref{ODE-Training-Task-ReWritten}. Notably, by forcing $\kappa_k\equiv 0$ for all $0\leq k\leq K-1$, the AL method degenerates the penalty approach established in the previous section.

Then, by calculus of variations \citep{liberzon2011calculus}, the adjoint variable $p_t^k$ is shown to satisfy the backward-in-time differential equations (see \autoref{appendix-Calculsu-Variations} for more details)
\begingroup
\renewcommand*{\arraystretch}{2.5}
\vspace{-0.22cm}
\begin{equation}
\begin{array}{l}
	\displaystyle dp_t^{k} = - p_t^{k}\frac{\partial f(x_t^{k},w^k_t)}{\partial x} dt \ \ \ \textnormal{on}\ [s_k,s_{k+1}),\\
	\displaystyle p^k_{s_{k+1}^-} = \left(1-\delta_{k,K-1}\right) \Bigg( \beta \frac{\partial \psi(\lambda_{k+1},x^k_{s_{k+1}^-})}{\partial x} + \kappa_{k+1} \Bigg) +\delta_{k,K-1} \, \frac{\partial \varphi(x^{k}_{s_{k+1}^-})}{\partial x},
\end{array}
	\label{Parallel-ODE-Adjoint-Equation-AL}
\end{equation}
\endgroup
for any $0\leq k\leq K-1$, while the update rule for control variables is given by
\begin{equation}
	w_t^k \leftarrow w_t^k - \eta p_t^{k} \frac{\partial f(x_t^k,w^k_t)}{\partial w} \ \ \ \textnormal{on} \ [s_k,s_{k+1}]
	\label{Parallel-ODE-Weights-Update-AL}
\end{equation}
for $0\leq k\leq K-1$, and the correction of auxiliary variables now takes on the form
\begin{equation}
	\lambda_0 \equiv x_0\ \ \  \textnormal{and}\ \ \  \lambda_k \leftarrow \lambda_k - \eta \left( \beta \frac{\partial \psi(\lambda_k,x^{k-1}_{s_{k}^-})}{\partial \lambda} + p^k_{s_k^+} - \kappa_k \right)\ \ \textnormal{for}\ 1\leq k\leq K-1.
	\label{Parallel-ODE-Slack-Variables-Update-AL}
\end{equation}
 
Similar to the analysis for penalty method, by choosing $\psi(\lambda,x) = \lVert \lambda-x \rVert_{\ell_2}^2$, formula \eqref{Parallel-ODE-Slack-Variables-Update-AL} implies that the constraint violations of augmented Lagrangian method satisfy
\begin{equation}
	  \lambda_0 - x^{-1}_{s_{0}^-} = 0\ \ \ \textnormal{and}\ \ \ \lambda_k - x^{k-1}_{s_{k}^-}  \approx \frac{1}{2\beta} (\kappa_k - p^k_{s_k^+})\ \ \textnormal{for}\ 1\leq k\leq K-1
	  \label{Constraint-Violation-AL}
\end{equation}
which offers two ways of improving the consistency constraint $x^{k-1}_{s_{k}^-}=x^{k}_{s_{k}^+}$: increasing $\beta$ or sending $\kappa_k\to p^k_{s_k^+}$, whereas the penalty method \eqref{Constraint-Violation-Penalty} provides only one option. As an immediate consequence, the ill-conditioning of quadratic penalty method can be avoided without increasing $\beta$ indefinitely. Under such circumstances, the update rule for explicit Lagrange multipliers can be deduced from \eqref{Constraint-Violation-AL}, namely,
\begin{equation}
	\kappa_0 \equiv 0\ \ \ \textnormal{and} \ \ \ \kappa_k \leftarrow \kappa_k - \frac{\eta}{2\beta} \big(\lambda_k -x^{k-1}_{s_k^-}\big)\ \ \textnormal{for}\ 1\leq k\leq K-1.
	\label{Parallel-ODE-Multiplier-Update-AL}
\end{equation}

To sum up, the essential ingredients of AL method for approximately solving the time-serial training problem \eqref{ODE-Training-Task} at each iteration includes
\begin{tcolorbox}[standard jigsaw,opacityback=0,]
\vspace{-0.52cm}					
\begin{empheq}{align*}
	\bullet \ \textnormal{decoupled local operations}\ \eqref{Parallel-ODE-State-Equation}, \eqref{Parallel-ODE-Adjoint-Equation-AL}, \eqref{Parallel-ODE-Weights-Update-AL} \ \qquad \ \bullet \ \textnormal{global communication}\ \eqref{Parallel-ODE-Slack-Variables-Update-AL}, \eqref{Parallel-ODE-Multiplier-Update-AL}
\end{empheq}
\end{tcolorbox}
\noindent which also parallelizes the time-serial iterative system \eqref{ResNet-ODE-KKT-System} and hence realizes the forward, backward, and update unlocking in the continuous-time sense. Moreover, by employing the AL method, the ill-conditioning of quadratic penalty method \eqref{Constraint-Violation-Penalty} can be lessened without increasing the penalty coefficient indefinitely. However, the introduction of explicit Lagrangian multipliers may require additional memory and communication overheads that need to be properly handled (see more discussion in the following section).

\section{Layer-Parallel Training} \label{section-parallel-training-algorithm}

According to the stable finite difference schemes discussed in Section \ref{section-ResNet} (or see reference \citep{gholami2019anode}), we are now ready to discretize the time-parallel training problem \eqref{Parallel-ODE-Training-Task} into local optimization processes and establish a non-intrusive layer-parallel training algorithm, \textit{i.e.}, the third path in \autoref{fig-big-picture}. Note that the external auxiliary variables can increase concurrency across all the building modules but would incur additional memory and communication overheads, which makes it quite challenging to cooperate with the indispensable data augmentation techniques \citep{tanner1987calculation}. Therefore, auxiliary-variable networks are proposed and deployed recursively to balance the storage overhead and the computation time, which enables the use of data augmentation for improving the performance of trained models.

\subsection{Decoupled Forward Pass and Backpropagation} \label{section-ParaTrain-no-AVNet}

Recall the partitioning of $[0,1]$ associated with the layer-serial training task \eqref{ResNet-Training-Task}, \textit{i.e.}, 
\begin{equation*}
	0 = t_0 < t_1 < \ldots < t_\ell = \ell\Delta t < t_{\ell+1} < \ldots < t_{L}=1,
\end{equation*}
its decomposed counterpart (see also Section \ref{section-parareal-terminal-control-FP}) is built by choosing a coarsening factor $n\in\mathbb{N}_+$ and extracting every $n$-th building block as depicted in \autoref{fig-multifidelity-forward-propagation}
\begin{equation*}
	0 = s_0  < \ldots < s_k = t_{nk} < s_{k+1} < \ldots < s_{K}=1,
\end{equation*}
namely, the network architecture within each coarse interval $[s_k,s_{k+1}]$ is constructed with a stack of $n$ residual blocks.

To be specific, $[s_k,s_{k+1}]$ can be uniformly divided into $n$ sub-intervals  
\begin{equation*}
	s_k = t_{kn} < t_{kn+1} < \cdots < t_{kn+n-1} < t_{kn+n} = s_{k+1},
\end{equation*}
for $0\leq k\leq K-1$,  and we have by \eqref{Parallel-ODE-State-Equation} that the feature flow now evolves according to
\begin{equation}
	X_{kn}^k =\lambda_k, \ \ \  X_{kn+m+1}^k = X_{kn+m}^k + F(X_{kn+m}^k,W^k_{kn+m})
	\label{Parallel-ResNet-Feature-Flow}
\end{equation}
where $0\leq m\leq n-1$, which can be executed in parallel across $K$ computing processors.

Next, by employing the particular numerical schemes \eqref{ResNet-Gradient-Propagation} that arise from the discrete-to-continuum analysis (see \autoref{appendix-Serial-Train-Time}) and equation \eqref{Parallel-ResNet-Feature-Flow}, the discretization of our adjoint equation \eqref{Parallel-ODE-Adjoint-Equation-AL} is given by the backward dynamic
\begingroup
\renewcommand*{\arraystretch}{2.5}
\vspace{-0.22cm}
\begin{equation}
\begin{array}{l}
	\displaystyle P_{kn+m}^k = P_{kn+m+1}^k  + P_{kn+m+1}^k \frac{\partial F(X_{kn+m}^k,W^k_{kn+m})}{\partial X} = P_{kn+m+1}^k \frac{\partial X^k_{kn+m+1}}{\partial X^k_{kn+m}},\\
	\displaystyle P_{kn+n}^k = (1-\delta_{k,K-1}) \left( \beta\frac{\partial \psi(\lambda_{k+1},X^k_{kn+n})}{\partial X^k_{kn+n}} + \kappa_{k+1} \right) + \delta_{k,K-1}  \frac{\partial \varphi(X_{kn+n}^{k})}{\partial X^k_{kn+n}}.
\end{array}
	\label{Parallel-ResNet-Gradient-Propagation-Equation}
\end{equation}
\endgroup
Or, equivalently, for any $0\leq k\leq K-1$ and $0\leq m\leq n$,  the discrete adjoint variable in \eqref{Parallel-ResNet-Gradient-Propagation-Equation} can be reformulated as
\begin{equation}
	P_{kn+m}^k = (1-\delta_{k,K-1}) \left( \beta\frac{\partial \psi(\lambda_{k+1},X^k_{kn+n})}{\partial X^k_{kn+m}} + \kappa_{k+1} \frac{\partial X^k_{kn+n}}{\partial X^k_{kn+m}} \right) + \delta_{k,K-1} \frac{\partial \varphi(X_{kn+n}^{k})}{\partial X^{k}_{kn+m}}
	\label{Parallel-ResNet-Gradient-Propagation}
\end{equation}
which captures the terminal and intermediate loss changes, \textit{i.e.}, the second- and first-term on the right-hand-side of \eqref{Parallel-ResNet-Gradient-Propagation}, with respect to the hidden activation layers for $k=K-1$ and $0\leq k\leq K-2$, respectively.

In contrast to the traditional method \citep{maday2002parareal,gunther2020layer} where the control updates are performed after finding all the discrete state and adjoint variables, we propose to conduct control updates \eqref{Parallel-ODE-Weights-Update-AL} right after the numerical solution of the adjoint equation \eqref{Parallel-ODE-Adjoint-Equation-AL} at each time slice. More specifically, by recalling \eqref{ODE-Weights-Update} and \eqref{ResNet-Parameter-Updates}, we have, for any $0\leq k\leq K-1$ and $0\leq m\leq n$, that the parameter updates associated with the $k$-th stacked modules can be deduced from \eqref{Parallel-ODE-Weights-Update-AL}, \eqref{Parallel-ResNet-Feature-Flow}, and \eqref{Parallel-ResNet-Gradient-Propagation}, \textit{i.e.},
\begin{equation}
\begin{split}
	W_{kn+m}^k \leftarrow \ & W_{kn+m}^k - \eta  P^k_{kn+m+1} \frac{\partial F(X^k_{kn+m},W^k_{kn+m})}{\partial W} \\
 = \ & W^k_{kn+m} - \eta \bigg(  (1-\delta_{k,K-1}) \bigg( \beta \frac{\partial \psi(\lambda_{k+1},X^k_{kn+n})}{\partial W^k_{kn+m}} + \kappa_{k+1} \frac{\partial X^k_{kn+n}}{\partial W^k_{kn+m}} \bigg) \\
	& \qquad \qquad \ \ \ \ \ + \delta_{k,K-1} \frac{\partial \varphi(X_{kn+n}^{k})}{\partial W^k_{kn+m}} \bigg)
	\label{Parallel-ResNet-Backpropagation}
\end{split}	
\end{equation}
which can not only be executed in parallel but also guarantee the accuracy of gradient information \citep{gholami2019anode}.

On the other hand, we have by \eqref{Parallel-ODE-Slack-Variables-Update-AL} and \eqref{Parallel-ResNet-Gradient-Propagation} that the correction of auxiliary variables satisfies $\lambda_0 \equiv x_0$ and
\begin{equation}
\begin{split}
	 \lambda_k \leftarrow \lambda_k - \eta & \bigg( \beta \frac{\partial \psi(\lambda_k,X^{k-1}_{kn})}{\partial \lambda} + (1-\delta_{k,K-1}) \bigg( \beta\frac{\partial \psi(\lambda_{k+1},X^k_{kn+n})}{\partial X^k_{kn}} + \kappa_{k+1}\frac{\partial X^k_{kn+n}}{\partial X^k_{kn}} \bigg) \\
	 & \ \ + \delta_{k,K-1} \frac{\partial \varphi(X_{kn+n}^{k})}{\partial X^{k}_{kn}} - \kappa_k \bigg) \ \ \textnormal{for}\ 1\leq k\leq K-1,
	\label{Parallel-ResNet-Slack-Variables-Update}
\end{split}	
\end{equation}
while the update rule for explicit Lagrangian multiplier $\kappa_k$ is given by
\begin{equation}
	\kappa_0 = 0\ \ \ \textnormal{and}\ \ \ \kappa_k \leftarrow \kappa_k - \frac{\eta}{2\beta}\left( \lambda_k - X^{k-1}_{kn} \right)\ \ \textnormal{for}\ 1\leq k\leq K-1,
	\label{Parallel-ResNet-Multiplier-Update}
\end{equation}
both of which require communication between adjacent modules. 

To sum up, the proposed layer-parallel training strategy for approximately solving the minimization problem \eqref{ResNet-Training-Task} can be formulated as
\begin{tcolorbox}[standard jigsaw,opacityback=0,]
\vspace{-0.52cm}					
\begin{empheq}{align*}
	\bullet \ \textnormal{local forward pass and backpropagation}\  \eqref{Parallel-ResNet-Feature-Flow}, \eqref{Parallel-ResNet-Backpropagation} \ \ \bullet \, \textnormal{global communication}\ \eqref{Parallel-ResNet-Slack-Variables-Update}, \eqref{Parallel-ResNet-Multiplier-Update}
\end{empheq}
\end{tcolorbox}
\noindent at each iteration, leading to a synchronous module parallelization, and hence realizes the forward, backward, and update unlocking. It is noteworthy that the forward locking is removed in a synchronous fashion, \textit{i.e.}, each of the split parts processes its input data from the same mini-batch (see \autoref{Table-Related-Works-Fig}). Moreover, by forcing all the explicit Lagrangian multipliers $\kappa_k$ to have null values, we arrive at the penalty approach discussed in section \ref{section-penalty-method}.

\subsection{Auxiliary-Variable Networks}

\begin{table}[h]
\setlength{\tabcolsep}{3pt}
\centering
\begin{tabular}{c c}
\toprule
Method & Forward Pass \\ 
\midrule
\parbox{2.5cm}{\centering ResNet \\ (layer-serial)} & \vspace*{0.05cm}\raisebox{-0.42\totalheight}{\includegraphics[width=0.85\textwidth]{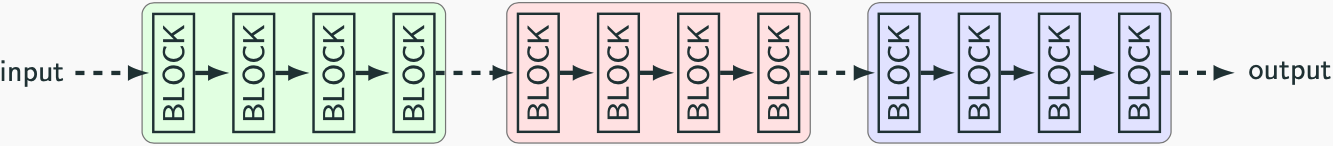}}  \\
\midrule
\parbox{2.5cm}{\centering ResNet \\ (layer-parallel)} & \vspace*{0.05cm}\raisebox{-0.42\totalheight}{\includegraphics[width=0.85\textwidth]{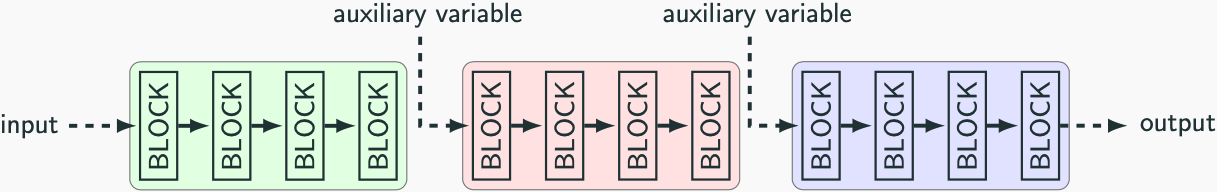}}  \\
\midrule
\parbox{2.5cm}{\centering ResNet \\ (layer-parallel \\ +AuxNet$\times1$)} & \vspace*{0.05cm}\raisebox{-0.42\totalheight}{\includegraphics[width=0.85\textwidth]{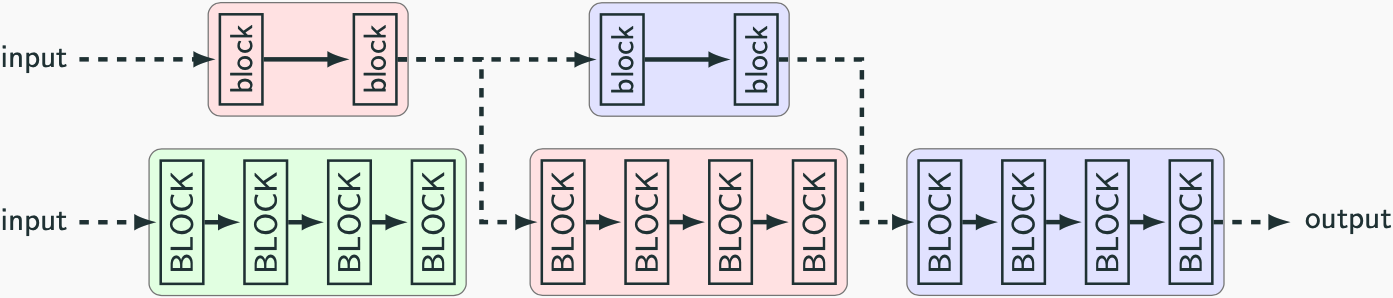}} \\
\midrule
\parbox{2.5cm}{\centering ResNet \\ (layer-parallel \\ +AuxNet$\times2$)} & \vspace*{0.05cm}\raisebox{-0.42\totalheight}{\includegraphics[width=0.85\textwidth]{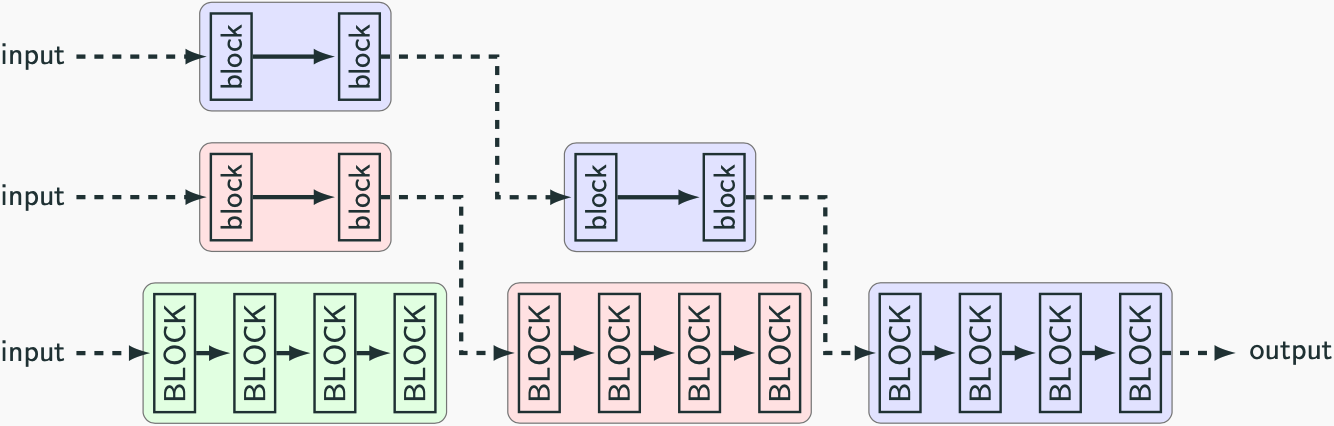}} \\
\bottomrule
\bottomrule
\vspace{-0.43cm} \\
\makecell{Notation \\ Description} & \raisebox{-0.42\totalheight}{\includegraphics[width=0.85\textwidth]{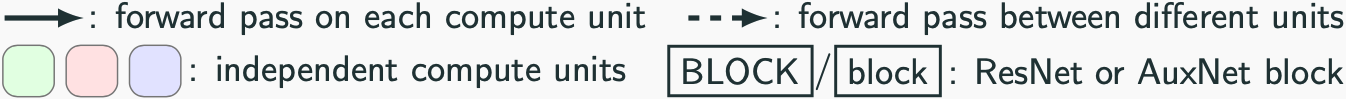}}  \\
\bottomrule
\end{tabular}
\caption{Decoupled forward pass through the use of auxiliary-variable networks, where the ResNet model is split and distributed across $K=3$ independent compute devices.}
\label{Table-Recursive-AuxNet-Fig}
\end{table}

Though the discrete dynamical systems in the previous section already offer us a fully decoupled training scheme, a key observation from \eqref{AL-functional} is that each input mini-batch requires to introduce a group of external auxiliary variables for fulfilling the classification task across multiple compute devices. When cooperating with the widely deployed data augmentation techniques\footnote{An effective regularization technique that reduces overfitting by artificially increasing the number of samples in the training set, \textit{e.g.}, randomly flipping, rotating, or shifting the input image.} \citep{tanner1987calculation} for enhancing the performance of deep learning models, the extra augmented images would incur prohibitive memory and communication overheads and therefore limit the performance for the exposed parallelism. Unfortunately, most of the existing auxiliary-variable methods fail to address this issue \citep{gotmare2018decoupling,choromanska2018beyond}, which often leads to a severe testing accuracy drop of trained models. For instance, the quadratic penalty method without using data augmentation \citep{gotmare2018decoupling} sees an accuracy drop of around $5\%$ in training decoupled ResNet-18 ($K=2$) on CIFAR-10 dataset, while a similar observation can be drawn from \autoref{table-DAR} when training ResNet-110 (implementation details can be found in \autoref{sectioin-experiment}).

\begin{table}[h]
\setlength{\tabcolsep}{3.3pt}
\centering
\begin{tabular}{cccccc}
\toprule
& \multicolumn{3}{c}{Training without Data Augmentation} & \multicolumn{2}{c}{Training with Data Augmentation} \\
\cmidrule(lr){2-4}
\cmidrule(lr){5-6}
& BP ($K\!=\!1$) & Penalty ($K\!=\!3$) & AL ($K\!=\!3$) & BP ($K\!=\!1$) & Penalty+AuxNet ($K\!=\!3$) \\
\midrule
Test Acc. & 86.4$\%$ & 84.5$\%$ & 85.2$\%$ & 93.7$\%$ & 92.5$\%$ \\
\bottomrule
\end{tabular}
\caption{Testing accuracy of (decoupled) ResNet-110 on CIFAR-10 dataset, in which the penalty and AL methods are implemented with or without data augmentation.}
\label{table-DAR}
\end{table}
\begin{figure}[h]
\centering
\includegraphics[width=0.8\columnwidth]{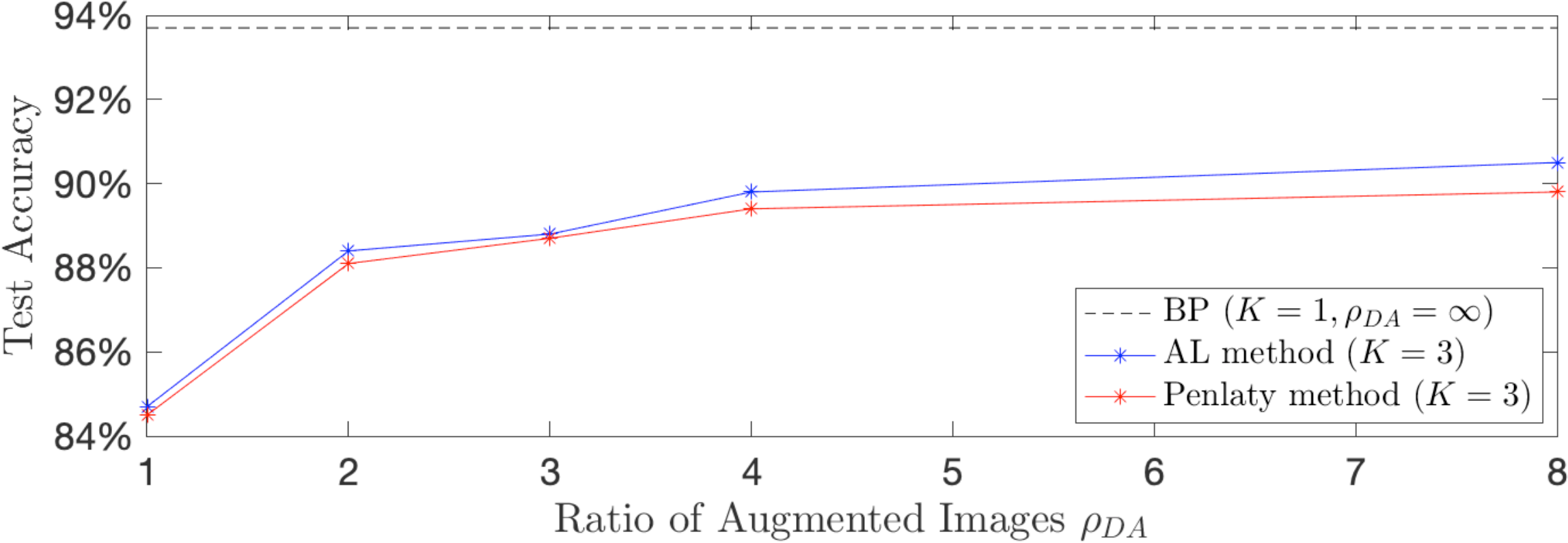}
\vspace{-0.2cm}
\caption{Testing accuracy of ResNet-110 on CIFAR-10 using data augmentation.}
\label{figure-DAR}
\end{figure}

To further justify our argument, we show in \autoref{figure-DAR} the test accuracy of decoupled ResNet-110 on CIFAR-10 dataset, where different ratios of augmented images $\rho_{\textnormal{DA}}$ (\textit{i.e.}, the number of synthetically modified images to that of real images) is utilized during training and $\rho_{\textnormal{DA}}=\infty$ denotes the data augmentation containing random operations. It can be observed that, as $\rho_{\textnormal{DA}}$ is increased, the accuracy gap between the traditional layer-serial training approach and the proposed layer-parallel training methods is tending to close. However, the memory requirements for storing all the augmented images and their corresponding auxiliary variables blow up even for moderate values of $\rho_{\textnormal{DA}}$, which could be unaffordable in practical scenarios.

To enable the use of random data augmentation during the layer-parallel training process, we propose a novel joint learning framework that allows trade-off between storage and recomputation of the external auxiliary variables. We can, for instance, take the example of penalty method. If not specified otherwise, the term auxiliary variables is used to only refer to $\{\lambda_k\}_{k=0}^{K-1}$ in what follows. Instead of writing the auxiliary variables directly to the CPU memory after each correction \eqref{Parallel-ResNet-Slack-Variables-Update} (see the second figure in \autoref{Table-Recursive-AuxNet-Fig}), an auxiliary-variable network is trained to learn the mapping between these auxiliary variables by replicating their values at the current iteration (see the third figure in \autoref{Table-Recursive-AuxNet-Fig}), that is, for $1\leq k\leq K-1$,
\begin{equation}
	\lambda_{k-1} \mapsto \lambda_{k} = \textnormal{AuxNet}^{(\lambda)}_{k-1}( \lambda_{k-1}; \theta_{k-1}^{(\lambda)} )
	\label{AV-Net-FP-lambda}
\end{equation}
or, equivalently, the AuxNet above is trained to solve the following minimization problem with pairs of input-output data obtained from \eqref{Parallel-ResNet-Slack-Variables-Update}
\begin{equation}
	   \operatorname*{arg\,min}_{\theta_{k-1}^{(\lambda)}} \big\lVert \lambda_{k} - \textnormal{AuxNet}^{(\lambda)}_{k-1}( \lambda_{k-1}; \theta_{k-1}^{(\lambda)} ) \big\rVert_{\ell_2}^2
	 \label{AV-Net-lambda}
\end{equation}
where $\theta_{k-1}^{(\lambda)}$ denotes the trainable parameters. Notably, in contrast to the classical ResNet \eqref{ResNet-Training-Task} that requires a large number of model parameters to guarantee its strong exploration capability for solving complex machine learning tasks, the auxiliary-variable network is constructed to mimic the approximate latent feature (or distilled knowledge) generated from the correction operations \eqref{Parallel-ResNet-Slack-Variables-Update}. This observation motivates us to employ a ResNet-like network with small capacity \citep{gou2021knowledge} for joint training with the layer-parallel training algorithms established in section \ref{section-ParaTrain-no-AVNet}, and the experimental results reported in \autoref{sectioin-experiment} validate the effectiveness of our proposed strategy.

As a result, the layer-parallel training strategy through the use of penalty method (\textit{i.e.}, forcing $\kappa_k\equiv 0$ during training) and auxiliary-variable networks can be formulated as
\begin{tcolorbox}[standard jigsaw,opacityback=0,]
\vspace{-0.52cm}					
\begin{empheq}{align*}
	\bullet \ \textnormal{forward pass of AuxNet}\, \eqref{AV-Net-FP-lambda} \ \ \ \bullet\, \textnormal{local forward pass and backpropagation}\, \eqref{Parallel-ResNet-Feature-Flow}, \eqref{Parallel-ResNet-Backpropagation} \\
	\bullet \ \textnormal{global communication}\ \eqref{Parallel-ResNet-Slack-Variables-Update} \ \ \ \bullet \, \textnormal{local knowledge distillation} \, \eqref{AV-Net-lambda} \qquad \qquad \  
\end{empheq}
\vspace{-0.6cm}	
\end{tcolorbox}
\noindent at each iteration, which achieves forward, backward, and update unlocking without storing the external auxiliary variables in CPU memory. To the best of our knowledge, this is the first study in the literature of training modern ResNets across real-world datasets that breaks the forward locking in a synchronous fashion (see also \autoref{Table-Related-Works-Fig} or \autoref{Table-Related-Works}).

It is noteworthy that although the explicit Lagrangian multiplier $\kappa_k$ has the same size as the auxiliary variable $\lambda_k$ for $1\leq k\leq K-1$, the forward dependency between $\kappa_{k-1}$ and $\kappa_k$ is not as straightforward as that of $\lambda_{k-1}$ and $\lambda_k$. As such, the remaining of this work will focus on the penalty method, and the design of AuxNet architecture for reproducing the explicit Lagrangian multipliers is left for future investigation.


\subsubsection{Non-Intrusive Implementation Details}

We summarize our results in \autoref{algorithm-parallel-training} for training modern ResNets across real-world datasets, which can be implemented in a non-intrusive way with respect to the existing network architectures\footnote{Trainable parameters in the input and output layers, \textit{i.e.}, $S$ and $T$ in \autoref{fig-ResNet-architecture}, can be automatically learned by coupling into the first and last workers respectively.}. Besides, we denote by $K=1$ the traditional layer-serial training method. Note that for a single iteration step, the computational time associated with the layer-serial ($K=1$) and layer-parallel ($K>1$) training methods are outlined as follows:
\begin{center}
\begin{tabular}{ccccc}
\toprule
Methods  & Data Load & Forward Pass & Backpropagation & Auxiliary Network \\ 
\midrule
Layer-Serial & $T_d$ & $T_f$ & $T_b$ & $0$ \\
\midrule
Layer-Parallel & $T_d$ & $\frac{1}{K}T_f$ & $\frac{1}{K}T_b + t_\psi $ & $t^{(\lambda)}_f + t^{(\lambda)}_b$ \\
\bottomrule
\end{tabular}
\end{center}
where $T_d$ denotes the time cost on data loader, $T_f$ $(T_b)$ the time cost of ResNet executing forward (backward) pass through the conventional layer-serial training approach,  $t_\psi$ the time of computing intermediate loss functions, $t^{(\lambda)}_f + t^{(\lambda)}_b$ the computation time executing forward pass and knowledge distillation of AuxNets. As a direct result, the speedup ratio per epoch\footnote{An epoch is a term indicating that neural network is trained with all the training data for one cycle.} can be expressed as
\begin{equation}
	\rho = \frac{\text{runtime per epoch (layer-serial training)}}{\text{runtime per epoch (layer-parallel training)}}  = \frac{1}{ \frac{1}{K} \frac{T_f+T_b}{T_d+T_f+T_b} + \frac{T_d+t_\psi + t^{(\lambda)}_f + t^{(\lambda)}_b}{T_d+T_f+T_b} }
	\label{speedup-ratio}
\end{equation}
where $T_d$, $T_f$, $T_b$, $t_\psi$, $t^{(\lambda)}_f$ and $t^{(\lambda)}_b$ are almost independent of the partition number $K$.

\begin{algorithm2e}[t]
\SetAlgoLined
\SetNlSty{texttt}{[\!}{\!]\,}
\SetAlgoNlRelativeSize{0}
\SetNlSkip{0em}
\tcp{Initialization.}
\nl split the baseline ResNet model into $K$ parts to deploy model parallelism\;
\nl initialize the trainable parameters of ResNets and auxiliary-variable networks\;
\nl schedule proper learning rates for ResNets and auxiliary-variable networks\; 
\nl use a proper penalty function with an increasing sequence of penalty coefficients\;

\tcp{Training Process.}
\nl \For{$j\leftarrow 1$ \KwTo $J$ (number of training epochs)}{
  \ForEach{mini-batch sampled from training dataset}{
  \tcp{forward pass to get auxiliary variables}
  \For{$k\leftarrow 0$ \KwTo $K-2$}{$\displaystyle \lambda_{k+1} = \textnormal{AuxNet}_{k}^{(\lambda)}(\lambda_k;\theta_k^{(\lambda)})$} 
  \tcp{decoupled forward-backward pass on multiple GPUs in parallel}
  \textbf{par}\For{$k\leftarrow 0$  \KwTo $K-1$}{
  \tcp{forward pass with auxiliary variable}
  \For{$m\leftarrow 0$ \KwTo $n-1$}{$\displaystyle X_{kn}^k =\lambda_{k}, \  X_{kn+m+1}^k = X_{kn+m}^k + F(X_{kn+m}^k,W_{kn+m})$}  
  \tcp{backpropagation with synthetic and classification losses}	
  \For{$m\leftarrow n$ \KwTo $0$}{$\displaystyle W^k_{kn+m} \leftarrow W^k_{kn+m} - \eta \bigg( (1-\delta) \beta \frac{\partial \psi(X^k_{kn+n},\lambda_{k+1})}{\partial W^k_{kn+m}} + \delta \frac{\partial \varphi(X_{kn+n}^{k})}{\partial W^k_{kn+m}} \bigg)$}
  }
  \tcp{communication across GPUs and local knowledge distillation }
    \textbf{par}\For{$k\leftarrow 1$ \KwTo $K-1$}{ 
    {$\displaystyle \lambda_{k} \leftarrow \lambda_{k} - \eta \bigg(  (1-\delta) \beta \frac{\partial \psi(X^k_{kn+n},\lambda_{k+1})}{\partial X^k_{kn}} + \delta \frac{\partial \varphi(X_{kn+n}^{k})}{\partial X^{k}_{kn}} + \beta \frac{\partial \psi(\lambda_{k},X^{k-1}_{kn})}{\partial \lambda} \bigg)$}\\	
	$\displaystyle \theta_{k-1}^{(\lambda)} \leftarrow \operatorname*{arg\,min} \big\lVert \lambda_{k} - \textnormal{AuxNet}^{(\lambda)}_{k-1}( \lambda_{k-1}; \theta_{k-1}^{(\lambda)} ) \big\rVert_{\ell_2}^2$\\
    }
 }}
\caption{Layer-parallel Training of ResNet with AuxNet (Penalty Method)}
\label{algorithm-parallel-training}
\end{algorithm2e}

Note that the capacity of AuxNets is much smaller than that of realistic ResNets \citep{he2016deep,he2016identity}, it is plausible to assume that $T_f+T_b> t_\psi + t^{(\lambda)}_f + t^{(\lambda)}_b$, which immediately shows speed-up over the traditional layer-serial training method by choosing a sufficient large value of $K$. Moreover, formula \eqref{speedup-ratio} also implies that the upper bound of speed-up ratio is given by
\begin{equation*}
	\rho < \frac{T_d + T_f+T_b}{T_d+t_\psi + t^{(\lambda)}_f + t^{(\lambda)}_b}.
\end{equation*}

\begin{remark}
We remark that a different definition of the speedup ratio has been applied by \citep{huo2018decoupled}. That is, by terminating the parallel training process once its testing accuracy is comparable to that of layer-serial training method, the speedup can then be defined as the ratio of serial execution time to the parallel execution time. Under such a circumstance, testing accuracy becomes the prerequisite for achieving speedup over the traditional method, and \citep{belilovsky2020decoupled} obtains higher speedup ratio by replacing the original loss function with artificially designed local loss functions during training.
\label{Remark-SpeedupRatio}
\end{remark}


\subsubsection{Recursive Auxiliary-Variable Networks}

As can be observed from the third figure in \autoref{Table-Recursive-AuxNet-Fig}, the time-consuming ResNet realizes forward unlocking with each of its split parts executing the forward pass in a synchronous fashion, however, its accompanied AuxNet still suffers from the issue of forward locking. 

To break the forward dependency between the external auxiliary variables, we propose to recursively employ the AuxNets, \textit{i.e.}, another AuxNet can be applied to remove the forward locking of its previous AuxNet and so forth (see the forth figure in \autoref{Table-Recursive-AuxNet-Fig} for example). As an immediate result, all the external auxiliary variables can be generated from the input mini-batch through a stack of AuxNets with different depths (which is referred to as ReAuxNet), that is, for $1\leq k\leq K-1$,
\begin{equation*}
\begin{split}
    	\lambda_{k} & = \textnormal{AuxNet}^{(\lambda)}_{k-1}( \textnormal{AuxNet}^{(\lambda)}_{k-2}(\cdots\textnormal{AuxNet}^{(\lambda)}_{0}(\lambda_0;\theta_0^{(\lambda)})\cdots;\theta_{k-2}^{(\lambda)}); \theta_{k-1}^{(\lambda)}) \\
    	& = \textnormal{ReAuxNet}^{(\lambda)}_{k}( X_0; \xi_{k}^{(\lambda)} )
\end{split}
\end{equation*}
where $\xi_{k}^{(\lambda)}$ represents the network parameters and $\lambda_{0} = X_0$. Accordingly, the decoupled forward pass \eqref{Parallel-ResNet-Feature-Flow} now takes on the form
\begin{equation}
	X_{kn}^k = \textnormal{ReAuxNet}^{(\lambda)}_{k}( X_0; \xi_{k}^{(\lambda)} ), \ \ X_{kn+m+1}^k = X_{kn+m}^k + F(X_{kn+m}^k,W^k_{kn+m}),
	\label{ReAuxNet-FP}
\end{equation}
for $0\leq k\leq K-1$ and $0\leq m\leq n-1$, while the update rule of ReAuxNet satisfies
\begin{equation}
\begin{split}
	 \xi_{k}^{(\lambda)} \leftarrow \xi_{k}^{(\lambda)} - \eta & \bigg( \beta \frac{\partial \psi(\lambda_k,X^{k-1}_{kn})}{\partial \lambda} + (1-\delta_{k,K-1}) \beta\frac{\partial \psi(\lambda_{k+1},X^k_{kn+n})}{\partial X^k_{kn}} \\
	 & \ \ + \delta_{k,K-1} \frac{\partial \varphi(X_{kn+n}^{k})}{\partial X^{k}_{kn}}  \bigg) \frac{\partial \textnormal{ReAuxNet}^{(\lambda)}_{k}( X_0; \xi_{k}^{(\lambda)} )}{\partial \xi_{k}^{(\lambda)}}
	\label{ReAuxNet-Distill}
\end{split}	
\end{equation}
for $1\leq k\leq K-1$. As a result, the layer-parallel training strategy through the use of penalty method (\textit{i.e.}, $\kappa_k\equiv 0$) is now given as below.
\begin{tcolorbox}[standard jigsaw,opacityback=0,]
\vspace{-0.52cm}					
\begin{empheq}{align*}
	\bullet\ \textnormal{local forward pass and backpropagation}\, \eqref{ReAuxNet-FP}, \eqref{Parallel-ResNet-Backpropagation} \ \ \ \bullet \, \textnormal{knowledge distillation} \, \eqref{ReAuxNet-Distill}
\end{empheq}
\end{tcolorbox}

Though the resulting layer-parallel training strategy is quite attractive due to its simplicity and ability to achieve forward unlocking in both the ResNet and AuxNets, the depth of ReAuxNet grows as the model is partitioned more finely. As a result, it requires more trainable parameters and thus increases the task difficulty for training large ResNets with fine-partitioned stages (see the forth figure in \autoref{Table-Recursive-AuxNet-Fig} for example). On the other hand, an interesting observation is that by forcing the first split parts of all ReAuxNets to have the same trainable parameters, the network structure depicted in the forth figure of \autoref{Table-Recursive-AuxNet-Fig} degenerates to that of the third one, which effectively reduces the total number of network parameters. Therefore, model compression through the use of shared weights is one potential solution to tackle this issue and is left for future investigation.

\section{Experiment}\label{sectioin-experiment}

To demonstrate the effectiveness and efficiency of our proposed strategy against the most commonly used BP approach and other compared methods, we conduct experiments using the benchmark ResNets \citep{he2016deep} and WideResNet \citep{zagoruyko2016wide} on CIFAR-10, CIFAR-100, and ImageNet datasets \citep{krizhevsky2009learning,krizhevsky2012imagenet}. All our experiments are implemented in Pytorch 1.4 \citep{paszke2019pytorch} using the multiprocessing library with NCCL backends. The ResNets, together with their accompanied AuxNets, are first split into $K$ pieces of stacked building blocks and then distributed on $K$ independent GPUs (Tesla-V100), where the standard data augmentation techniques, \textit{e.g.}, random crop and flip, are employed during the entire training procedure. 

As a preliminary study, we consider the penalty method presented in \autoref{algorithm-parallel-training} and report the training loss\footnote{The training loss refers to the error obtained through a full layer-serial forward pass of trained network.}, testing accuracy, constraint violation, and speedup ratio. It is noteworthy that in contrast to the regime discussed in remark \ref{Remark-SpeedupRatio}, our definition of speedup ratio is given by \eqref{speedup-ratio} and is widely adopted in the parallel computing community \citep{li2020training}.

\subsection{ResNet-110 on CIFAR-10 Dataset}

To begin with, we consider the benchmark ResNet-110 \citep{he2016identity} on CIFAR-10 dataset \citep{krizhevsky2009learning}, where the network architecture is constructed as that described in \citep{he2016deep} and the total number of trainable parameters is around 1.7 million. Models are trained on the 50k training images with a batch size of 128 for 300 epochs, which are then evaluated on the 10k testing images. The initial learning rate of stochastic gradient descent optimizer is set to $0.1$, and then decreases according to a cosine schedule \citep{loshchilov2016sgdr}. The auxiliary variables are perturbed by the Gaussian noise with small variance to prevent from overfitting. If not specified, the penalty coefficient is set to $\beta=10^5$ in what follows. 

\subsubsection{Auxiliary-Variable Network Design} 
Recall that the AuxNet is designed to mimic the feature maps that generate from the baseline ResNet-110 of different depths, hence a straightforward choice is to employ the original ResNet-110 as our AuxNet. As can be seen from \autoref{table-cifar10-smallNet}, the testing accuracy of such a setting is almost the same as the standard BP method, but the additional communication overheads would hamper the speedup ratio. 

\begin{table}[h]
\centering
\begin{tabular}{cccccc}
\toprule
& \multirow{2}{*}[-0.27em]{ResNet-110} & \multicolumn{4}{c}{AuxNet for Decoupled ResNet-110 $(K=3)$} \\
\cmidrule(lr){3-6}
& & ResNet-110 & ResNet-54 & ResNet-32 & ResNet-20 \\
\midrule
Test Acc. & 93.7$\%$ & 93.5$\%$ & 93.0$\%$ & 92.8$\%$ & 92.5$\%$ \\
\midrule
Speedup & - & 0.71 & 1.10 & 1.30 & 1.45 \\
\bottomrule
\end{tabular}
\caption{Testing accuracy and speedup of decoupled ResNet-110 on CIFAR-10, where the AuxNet with different capacities is employed for the joint training framework.}
\label{table-cifar10-smallNet}
\end{table}
\begin{figure}[h]
\centering
\includegraphics[width=.32\textwidth]{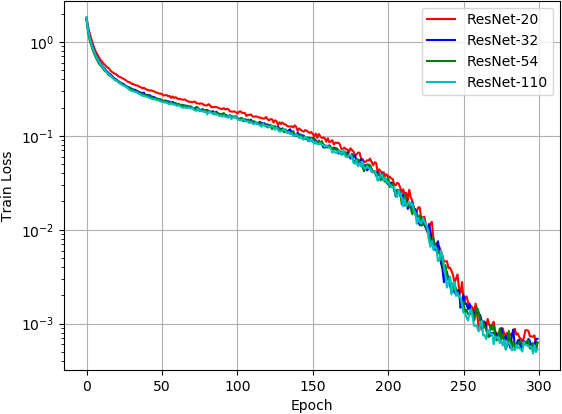}\hfill
\includegraphics[width=.32\textwidth]{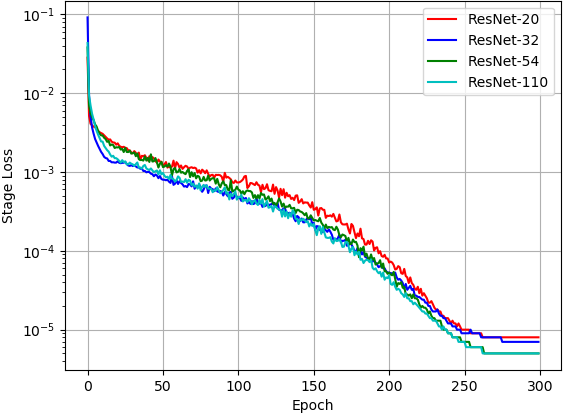}\hfill
\includegraphics[width=.32\textwidth]{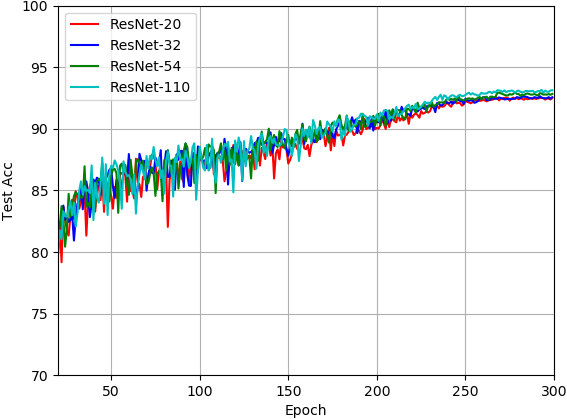}
\caption{Learning curves of training loss, constraint violation, and test accuracy for decoupled ResNet-110 ($K=3$) on CIFAR-10, where different AuxNets are employed.}
\label{Fig-Learning-Curves-resnet110}
\end{figure}

On the other hand, the AuxNet is not proposed to solve the classification task, which allows us to reduce its network capacity and further accelerates the training process. As shown in \autoref{table-cifar10-smallNet}, the speedup ratio increases as the model capacity of AuxNet declines, while the testing accuracy still remains comparable to the standard BP method (see also \autoref{Fig-Learning-Curves-resnet110}). To further validate the effectiveness of low-capacity auxiliary-variable network for fulfilling the joint learning task, we employ the ResNet-20 as the AuxNet and conduct experiments using ResNet-32, ResNet-54, and ResNet-110 on CIFAR-10 dataset. The experimental results are reported in \autoref{table-cifar10-BigNet}, which shows that our method can boost the training speed of various ResNets while maintain a comparable testing accuracy.

\begin{table}[h]
\centering
\begin{tabular}{ccccc}
\toprule
& & ResNet-110 & ResNet-54 & ResNet-32 \\
\midrule
Layer-Serial & Test Acc. & 93.7$\%$ & 93.3$\%$ & 92.6$\%$ \\
\midrule
\multirow{2}{*}[-0.27em]{Layer-Parallel} & Test Acc. & 92.5$\%$ & 92.0$\%$ & 91.3$\%$ \\
\cmidrule(lr){2-5}
 & Speedup & 1.45 & 1.18 & 1.01 \\
\bottomrule
\end{tabular}
\caption{Testing accuracy and speedup of different ResNets on CIFAR-10, where ResNet- 20 is employed as the AuxNet for decoupled training ($K=3$).}
\label{table-cifar10-BigNet}
\end{table}

\subsubsection{Stage Number and Penalty Coefficient} 

To further accelerate the network training procedure, we divide the baseline ResNet-110 into $K=4$ stages and summarize the experimental results in \autoref{table-cifar10-stage}. As the number of decoupled stages grows, the speedup ratio increases but the testing performance of trained model degenerates. This is because that more auxiliary variables and communication overheads are involved as the model is partitioned more finely, making the learning task harder to solve while inducing additional constraint violations. In other words, the increasing number of decoupled stages generally companies with a slight drop in performance. Under such circumstances, a wise strategy for scheduling the penalty coefficient is of crucial importance in order to force the minima of \autoref{algorithm-parallel-training} close to the feasible region of our original problem \eqref{ResNet-Training-Task}.

\begin{table}[h]
\centering
\begin{tabular}{ccccc}
\toprule
& $K=1$ & $K=2$ & $K=3$ & $K=4$ \\
\midrule
Test Acc. & 93.7$\%$ & 92.8$\%$ & 92.5$\%$ & 91.8$\%$ \\
\midrule
Speedup & - & 1.26 & 1.45 & 1.68 \\
\bottomrule
\end{tabular}
\caption{Testing accuracy and speedup of ResNet-110 with different values ($K$) of parallel stages on CIFAR-10, where the ResNet-20 is employed as AuxNet.}
\label{table-cifar10-stage}
\end{table}

\begin{figure}[h]
\centering
\includegraphics[width=.32\textwidth]{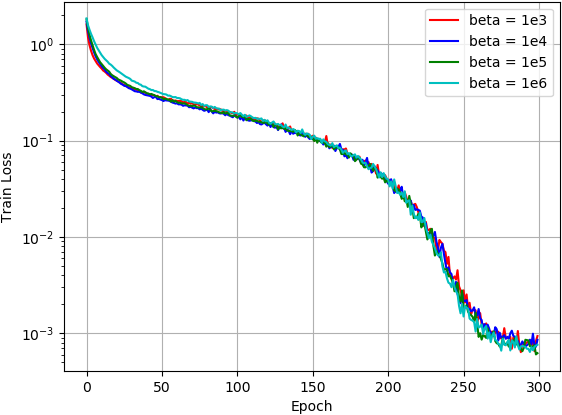}\hfill
\includegraphics[width=.32\textwidth]{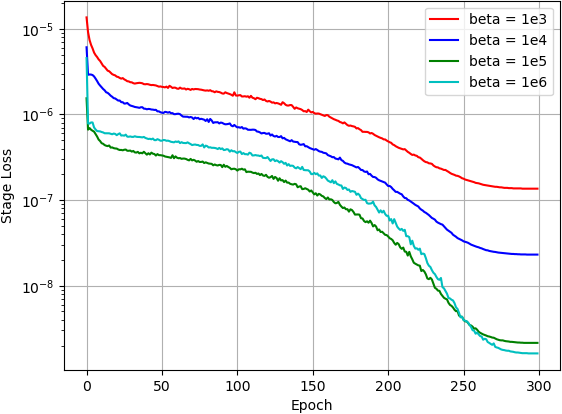}\hfill
\includegraphics[width=.32\textwidth]{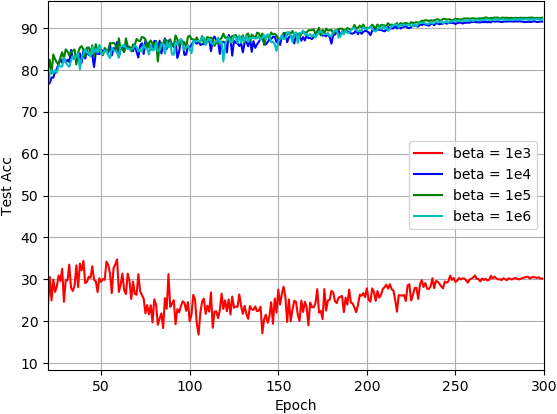}
\caption{Learning curves of the training loss, penalty loss, and testing accuracy for decoupled ResNet-110 on CIFAR-10 $(K=3)$, where different values of $\beta$ are used.}
\label{Fig-Learning-Curves-beta}
\end{figure}

Next, we conduct experiments for the decoupled training of ResNet-110 with different values of penalty coefficient. As can be seen from \autoref{Fig-Learning-Curves-beta}, even with the use of different penalty coefficients, the learning curves of the training loss have similar behaviours, \textit{i.e.}, all the optimization problems can be well-solved. However, training with a small penalty coefficient $\beta=10^3$ over-explores the infeasible region, which prematurely converges into an infeasible solution that leads to performance degradation (see \autoref{table-cifar10-beta}). On the contrary, in the case of a very large penalty coefficient, our proposed method may converge just after the feasibility is guaranteed, and hence fails to explore the infeasible region properly. To conclude, the value of penalty coefficient determines the superiority between the classification term and penalty term in the objective loss function \eqref{Parallel-ODE-Training-Task}, and a penalty coefficient that is too small or too large will result in performance degradation.

\begin{table}[h]
\centering
\begin{tabular}{ccccc}
\toprule
& $\beta=10^3$ & $\beta=10^4$ & $\beta=10^5$ & $\beta=10^6$ \\
\midrule
Test Acc. & 37.8$\%$ & 91.7$\%$ & 92.5$\%$ & 92.3$\%$ \\
\bottomrule
\end{tabular}
\caption{Testing accuracy of decoupled ResNet-110 ($K=3$) using different penalty coefficients on CIFAR-10, where the ResNet-20 is employed as the AuxNet.}
\label{table-cifar10-beta}
\end{table}

\subsubsection{Comparison to Other Methods} 

We then compare our joint learning approach with other representative methods from literature for solving the image classification task \eqref{ResNet-Training-Task} across CIFAR-10 dataset. Recall from \autoref{Table-Related-Works} or \autoref{Table-Related-Works-Fig} that the algorithmic locking issues of BP are not addressed by neither PipeDream nor Gpipe while the loss function of DGL methods is not consistent with the original learning task, the comparison is therefore made against the straight-forward implementation of penalty method without using data augmentation \citep{gotmare2018decoupling}, FR \citep{huo2018training}, DDG \citep{huo2018decoupled} and BP \citep{he2016deep}. We follow exactly the same setup from \citep{huo2018decoupled,huo2018training} and report the experimental results in \autoref{table-comparison-ResNet110-CIFAR10}. It is noteworthy that in contrast to the definition of speedup ratio applied in \citep{huo2018decoupled}, we adopt the more commonly used definition \eqref{speedup-ratio} for measuring the speed gain of parallel processing. Under such a circumstance, both DDG and FR show no speedup over the traditional layer-serial method (see \autoref{table-comparison-ResNet110-CIFAR10}), which is caused by the time-consuming communication overheads per epoch during training. On the other hand, it can be concluded from \autoref{table-comparison-ResNet110-CIFAR10} that enabling the use of data augmentation at training time can significantly improves the performance of naive penalty method, which validates the effectiveness and efficiency of AuxNet for fulfilling the joint learning task. Besides, our method is also comparable to that of FR and DDG in terms of testing accuracy. 

\begin{table}[h]
\centering
\begin{tabular}{cccccc}
\toprule
& BP & Penalty & DDG & FR & Our Method \\
\midrule
Test Acc. & 93.7$\%$ & 84.5$\%$ & 93.4$\%$ & 93.8$\%$ & 92.5$\%$ \\
\midrule
Speedup & - & 1.56 & 0.81 & 0.63 & 1.45 \\
\bottomrule
\end{tabular}
\caption{Testing accuracy and speedup of various ResNet training strategies on CIFAR-10 dataset, where the baseline ResNet-110 is decoupled into $K=3$ pieces.}
\label{table-comparison-ResNet110-CIFAR10}
\end{table}

\subsection{Large-Scale Experiments}

Note that, in contrast to the conventional auxiliary-variable methods, a key advantage of our strategy is that the algorithmic locking problems can be removed without explicitly storing the external auxiliary variables. This allows us to conduct experiments on large-scale datasets, and the results further demostrate the effectiveness of our methods.  

More specifically, we first work with WideResNet-40-10 \citep{zagoruyko2016wide} for CIFAR-100 dataset \citep{krizhevsky2009learning}, where the feature map is roughly 10 times larger than that of ResNet-110 on CIFAR-10. Accordingly, the size of auxiliary variables increases significantly as the neural network gets wider, resulting in a very large-scale optimization problem that is challenging to solve. The training starts with a learning rate of $0.1$, which then decays according to the cosine schedule \citep{loshchilov2016sgdr} for 200 epochs. The initial penalty coefficient is set to $10^6$, and is multiplied by $10$ after 50 epochs. As can be seen from \autoref{table-cifar100-AV}, our joint learning framework still works well under such a challenging scenario, achieving notable speedup with an acceptable drop of testing accuracy.

\begin{table}[h]
\centering
\begin{tabular}{ccccc}
\toprule
& \multirow{2}{*}[-0.27em]{WideResNet-40-10} & \multicolumn{3}{c}{AuxNet for Decoupled WideResNet-40-10} \\
\cmidrule(lr){3-5}
& & WideResNet-28-10 & & WideResNet-16-10  \\
\midrule
Test Acc. & 81.4$\%$ & 80.1$\%$ & & 79.0$\%$ \\
\midrule
Speedup & - & 1.16 & & 1.53  \\
\bottomrule
\end{tabular}
\caption{Testing accuracy and speedup of decoupled WideResNet-40-10 on CIFAR-100, in which the AuxNet with different capacities is employed and $K=3$.}
\label{table-cifar100-AV}
\end{table}

Next, we consider the more challenging ImageNet dataset \citep{krizhevsky2012imagenet} and compare the proposed training approach with other representative methods. More specifically, the ResNet-50 \citep{he2016identity} is chosen as the baseline model, together with ResNet-18 being employed as our AuxNet. All models are split across $K=2$ independent GPUs, and the training terminates after 90 epochs. The initial learning rate is $0.1$ and is divided by 10 every 30 epochs, while the penalty coefficient is set to $\beta=10^8$. The experimental results are displayed in \autoref{table-imagenet}, where all methods except the FR achieve speedup over the standard BP but suffer more or less from the drop of accuracy. This is because that the speedup ratio is defined by \eqref{speedup-ratio} rather than that in \citep{huo2018decoupled,huo2018training}. Notably, our method outperforms the other parallel training strategies in terms of test accuracy while the speedup is comparable to that of Sync-DGL.

\begin{table}[h]
\centering
\begin{tabular}{ccccc}
\toprule
& BP & FR & Sync-DGL & Our Method \\
\midrule
Test Acc. & 76.5$\%$ & 74.4$\%$ & 73.5$\%$ & 74.6$\%$ \\
\midrule
Speedup & - & 0.70 & 1.44 & 1.40  \\
\bottomrule
\end{tabular}
\caption{Testing accuracy and speedup of various methods on ImageNet dataset, in which the ResNet-50 is decoupled into $K=2$ pieces and ResNet-18 is used as AuxNet.}
\label{table-imagenet}
\end{table}

\section{Concluding Remarks}

In this paper, a novel joint learning framework is proposed for the distributed training of modern ResNets across real-world datasets, which fully decouples the conventional forward-backward training process through the employment of low-capacity auxiliary-variable networks. The similarity of training ResNets to the terminal control of neural ODEs motivates us to first utilize the penalty and AL methods for breaking the algorithmic locking in the continuous-time sense, and then to apply a consistent discretization scheme to achieve forward, backward, and update unlocking. Moreover, by trading off storage and recomputation of the external auxiliary variables, the proposed AuxNet enables the use of data augmentation during training, which is of great significance for improving the performance of trained models but not addressed by the existing auxiliary-variable methods. Experimental results are reported to validate the effectiveness and efficiency of our method. For future work, we plan to explore the viability of recursive AuxNets and design efficient AuxNets for reproducing the explicit Lagrangian multipliers.




\newpage

\begin{appendices}
\noindent\textbf{Appendix}
\vspace{-0.2cm}
\section{Layer-Serial Training} \label{appendix-Serial-Train} 
Note that by defining $\displaystyle P_{\ell+1} = \frac{\partial \varphi(X_L)}{\partial X_{\ell+1}}$ for $0\leq\ell\leq L-1$, scheme \eqref{ResNet-BackPropagation} can be rewritten as
\begin{equation}
	W_\ell \leftarrow W_\ell - \eta P_{\ell+1}\frac{\partial F(X_\ell,W_\ell)}{\partial W}, \qquad 0\leq \ell \leq L-1,
	\label{ResNet-Parameter-Updates}
\end{equation}
where ${\{P_{\ell+1}\}_{\ell=0}^{L-1}}$ satisfy a backward dynamic that captures the loss changes with respect to the hidden activations, \textit{i.e.},
\begin{equation}
	P_\ell  = P_{\ell+1} + P_{\ell+1} \frac{\partial F(X_\ell,W_\ell)}{\partial X}, \qquad  P_L = \frac{\partial \varphi(X_L)}{\partial X_L}. 
	\label{ResNet-Gradient-Propagation}
\end{equation}
Such a sequential propagation of error gradient is also known as the backward locking \citep{jaderberg2017decoupled}, preventing all layers of the network from updating until their dependent layers have executed the backward computation \eqref{ResNet-Gradient-Propagation}.

Consequently, the layer-serial BP algorithm \eqref{ResNet-BackPropagation} is handled by formulae \eqref{ResNet-Gradient-Propagation} and \eqref{ResNet-Parameter-Updates}. Or, to put it differently, the training process of ResNets at each epoch is performed through the repeated execution of
\begin{tcolorbox}[standard jigsaw,opacityback=0,]
\vspace{-0.52cm}					
\begin{empheq}{align*}
	\bullet \ \textnormal{forward pass in \eqref{ResNet-Training-Task}} \ \bullet \, \textnormal{backward gradient propagation}\ \eqref{ResNet-Gradient-Propagation} \ \bullet \, \textnormal{parameter updates}\ \eqref{ResNet-Parameter-Updates}
\end{empheq}
\end{tcolorbox}
\noindent which can be time-consuming as it is common to see neural networks with hundreds or even thousands of layers.

\section{Time-Serial Training} \label{appendix-Serial-Train-Time} 
By introducing the Lagrange functional with multiplier $p_t$ \citep{nocedal2006numerical}, solving the constrained optimization problem \eqref{ODE-Training-Task} is equivalent to finding saddle points of the following Lagrange functional without constraints\footnote{For notational simplicity, $\frac{dx_t}{dt}$ and $\dot{x}_t$ are used to denote the time derivative of $x_t$ throughout this work.}
\begin{equation*}
\begin{split}
	\mathcal{L}(x_t,w_t,p_t) & = \varphi(x_1) + \int_0^1 p_t\left(f(x_t,w_t)-\dot{x}_t\right) dt \\
	& =  \varphi(x_1) - p_1x_1 + p_0x_0 + \int_0^1 p_tf(x_t,w_t) + \dot{p}_tx_t\,dt.
\end{split}
\end{equation*}
and the variation in functional $\mathcal{L}(x_t,w_t,p_t)$ corresponding to a variation $\delta w$ in the control variable $w_t$ takes on the form \citep{liberzon2011calculus}
\begin{equation*}
	\delta\mathcal{L} = \bigg[ \frac{\partial \varphi(x_1)}{\partial x} - p_1 \bigg] \delta x + \int_0^1 \bigg( p_t\frac{\partial f(x_t,w_t)}{\partial x} + \dot{p}_t \bigg)\delta x + \bigg( p_t\frac{\partial f(x_t,w_t)}{\partial w} \bigg) \delta w \, dt,
\end{equation*}
which leads to the necessary conditions for $w_t=w_t^*$ to be the extremal of $\mathcal{L}(x_t,w_t,p_t)$, \textit{i.e.},
\begin{align*}
    & dx_t^* = f(x_t^*,w_t^*)dt, & & x_0^*=S(y), & & (\textnormal{state equation}) \\
    & dp_t^* = -p_t^* \frac{\partial f(x_t^*,w_t^*)}{\partial x}dt, & & p_1^*=\frac{\partial \varphi(x_1^*)}{\partial x},  & & (\textnormal{adjoint equation}) \\
    & p_t^* \frac{\partial f(x_t^*,w_t^*)}{\partial w} = 0, & & \textnormal{for}\ 0\leq t\leq 1. & & (\textnormal{optimality condition})  
\end{align*}
However, directly solving this optimality system is computationally infeasible, a gradient-based iterative approach with step size $\eta>0$ is typically used, \textit{e.g.},
\begin{subequations}\label{ResNet-ODE-KKT-System}
  \begin{align}
    & dx_t = f(x_t,w_t)dt, & & x_0=S(y), & & (\textnormal{forward pass})  \\ 
    & dp_t = -p_t \frac{\partial f(x_t,w_t)}{\partial x}dt, & & p_1=\frac{\partial \varphi(x_1)}{\partial x}, & &  (\textnormal{backward gradient propagation})  \\ 
	& w_t \leftarrow w_t - \eta p_t \frac{\partial f(x_t,w_t)}{\partial w}, & &  \textnormal{for}\ 0\leq t\leq 1, & &  (\textnormal{parameter updates}) 
  \end{align}
\end{subequations}
which is consistent with the layer-serial training process through forward-backward propagation, \textit{i.e.}, \eqref{ResNet-Training-Task}, \eqref{ResNet-Gradient-Propagation} and \eqref{ResNet-Parameter-Updates}, by taking the limit as $L\to\infty$ \citep{li2017maximum}. In other words, the classic BP scheme \eqref{ResNet-BackPropagation} for solving problem \eqref{ResNet-Training-Task} can be recovered from \eqref{ODE-Adjoint-Equation} and \eqref{ODE-Weights-Update} after employing the stable discretization schemes \eqref{ResNet-Gradient-Propagation} and \eqref{ResNet-Parameter-Updates}.

\section{Time-Parallel Training (Augmented Lagrangian Method)} \label{appendix-Calculsu-Variations}

Recall that the augmented Lagrangian functional 
\begin{equation*}
\begin{split}
	& \mathcal{L}_{\textnormal{AL}}( x_t^k, p_t^k, w^k_t, \lambda_k, \kappa_k ) = \mathcal{L}_{\textnormal{P}}( x_t^k, p_t^k, w^k_t, \lambda_k ) - \sum_{k=0}^{K-1} \kappa_k( \lambda_k - x^{k-1}_{s_{k}^-} ) \\
 	= \ & \varphi(x^{K-1}_{s_K^-})  + \sum_{k=0}^{K-1} \bigg( \beta \psi(\lambda_k,x^{k-1}_{s_{k}^-}) - \kappa_k( \lambda_k - x^{k-1}_{s_{k}^-} ) + \int_{s_k}^{s_{k+1}} p^k_t \big( f(x_t^k,w^k_t) - \dot{x}_t^k \big)dt \bigg) \\
 	= \ & \varphi(x^{K-1}_{s_K^-})  + \sum_{k=0}^{K-1} \bigg( \beta \psi(\lambda_k,x^{k-1}_{s_{k}^-}) - \kappa_k( \lambda_k - x^{k-1}_{s_{k}^-} ) + \int_{s_k}^{s_{k+1}} p^k_t f(x_t^k,w^k_t) + \dot{p}_t^k x_t^k\,dt \\
 	& \qquad \qquad \qquad \ \ \ \ \ - p^k_{s_{k+1}^-}x^k_{s_{k+1}^-} + p^k_{s_{k}^+}\lambda_k  \bigg)
\end{split}	
\end{equation*}
can be decomposed as parts involving $x_t^{K-1}$ and $\{x_t^k\}_{k=0}^{K-2}$, \textit{i.e.},
\begin{equation*}
	\varphi(x^{K-1}_{s_K^-}) - p_{s_K^-}^{K-1}x_{s_K^-}^{K-1} + p_{s_{K-1}^+}^{K-1}\lambda_{K-1} + \int_{s_{K-1}}^{s_K} p_t^{K-1}f(x_t^{K-1},w^{K-1}_t) + \dot{p}_t^{K-1}x_t^{K-1}\,dt,
\end{equation*}
and 
\begin{equation*}
	\sum_{k=0}^{K-2} \beta \psi(\lambda_{k+1},x^k_{s_{k+1}^-}) + (\kappa_{k+1} - p^k_{s_{k+1}^-})x^k_{s_{k+1}^-} + p^k_{s_{k}^+}\lambda_k + \int_{s_k}^{s_{k+1}} p_t^{k}f(x_t^{k},w^k_t) + \dot{p}_t^{k}x_t^{k}\,dt -\kappa_{k+1}\lambda_{k+1} 
\end{equation*}
respectively. Then the variation in $\mathcal{L}_{\textnormal{AL}}( x_t^k, p_t^k, w^k_t, \lambda_k,\kappa_k )$ corresponding to a variation $\delta w_t^k$ in control $w^k_t$ takes on the form
\begin{equation*}
\begin{split}
	\delta\mathcal{L}_{\textnormal{AL}} & = \Bigg( \frac{\partial \varphi(x^{K-1}_{s_K^-})}{\partial x} - p_{s_K^-}^{K-1} \Bigg) \delta x^{K-1} + \int_{s_{K-1}}^{s_K} \Bigg( p_t^{K-1}\frac{\partial f(x_t^{K-1},w^{K-1}_t)}{\partial x} + \dot{p}_t^{K-1} \Bigg)\delta x^{K-1} \, dt \\
	& + \sum_{k=0}^{K-2} \Bigg( \beta \frac{\partial \psi(\lambda_{k+1},x^k_{s_{k+1}^-})}{\partial x} + \kappa_{k+1} - p_{s_{k+1}^-}^k \Bigg) \delta x^k + \int_{s_k}^{s_{k+1}} \bigg( p_t^k\frac{\partial f(x_t^k,w^k_t)}{\partial x} + \dot{p}_t^k \bigg)\delta x^k \, dt,
\end{split}	
\end{equation*} 
which implies that the adjoint (or co-state) variable $p_t^k$ satisfies the backward differential equations \eqref{Parallel-ODE-Adjoint-Equation-AL} \citep{liberzon2011calculus}, \textit{i.e.}, for any $0\leq k\leq K-1$,
\begingroup
\renewcommand*{\arraystretch}{2.5}
\vspace{-0.22cm}
\begin{equation*}
\begin{array}{l}
	\displaystyle dp_t^{k} = - p_t^{k}\frac{\partial f(x_t^{k},w^k_t)}{\partial x} dt \ \ \textnormal{on}\ [s_k,s_{k+1}),\\
	\displaystyle p^k_{s_{k+1}^-} = \left(1-\delta_{k,K-1}\right) \left( \beta \frac{\partial \psi(\lambda_{k+1},x^k_{s_{k+1}^-})}{\partial x} + \kappa_{k+1} \right) +\delta_{k,K-1} \, \frac{\partial \varphi(x^{k}_{s_{k+1}^-})}{\partial x}.
\end{array}
\end{equation*}
\endgroup

On the other hand, it can be easily deduced that the control updates satisfy
\begin{equation*}
	w_t^k \leftarrow w_t^k - \eta p_t^{k} \frac{\partial f(x_t^k,w^k_t)}{\partial w} \ \ \ \textnormal{on} \ [s_k,s_{k+1}]
\end{equation*}
for $0\leq k\leq K-1$, and the correction of auxiliary variables now takes on the form
\begin{equation*}
	\lambda_0 \equiv x_0 \ \ \ \textnormal{and} \ \ \ \lambda_k \leftarrow \lambda_k - \eta \Bigg( \beta \frac{\partial \psi(\lambda_k,x^{k-1}_{s_{k}^-})}{\partial \lambda} + p^k_{s_k^+} - \kappa_k \Bigg) \ \ \textnormal{for} \ 1\leq k\leq K-1.
\end{equation*}

Obviously, by forcing the Lagrangian multiplier $\kappa_k\equiv 0$ for all $0\leq k\leq K-1$, the augmented Lagrangian method degenerates to the standard penalty approach. 

\end{appendices}

\vskip 0.2in
\bibliography{references}

\end{document}